\pdfoutput=1
\documentclass[]{amazonpreprint}

\usepackage{latexsym}
\usepackage[T1]{fontenc}
\usepackage[utf8]{inputenc}
\usepackage{microtype}
\usepackage{inconsolata}
\usepackage{graphicx}
\usepackage{booktabs}
\usepackage{multirow}
\usepackage{tabularx}
\usepackage{makecell}
\usepackage{array}
\usepackage{amsmath}
\usepackage{amssymb}
\usepackage{xspace}
\usepackage{enumitem}
\usepackage{url}
\usepackage{xcolor}
\usepackage{hyperref}
\usepackage{tcolorbox}
\usepackage{caption}
\usepackage{listings}
\usepackage{wrapfig}
\usepackage{adjustbox}
\usepackage{afterpage}

\definecolor{lightgreen}{rgb}{0.88, 1, 0.88}
\definecolor{lightpurple}{RGB}{180, 150, 255}
\definecolor{lightred}{RGB}{255, 204, 203}
\definecolor{lightblue}{RGB}{173, 216, 230}
\tcbuselibrary{skins}
\newtcolorbox{promptbox}[1][]{
  enhanced,
  colback=gray!8,
  colframe=black,
  boxrule=0.6pt,
  arc=3mm,
  left=3mm,right=3mm,top=2mm,bottom=2mm,
  drop shadow=black!30,
  title=#1,
  fonttitle=\bfseries,
}

\definecolor{takeawayfill}{gray}{0.93}
\definecolor{takeawayrule}{gray}{0.55}
\newtcolorbox{takeawaybox}{
  colback=takeawayfill, colframe=takeawayrule,
  boxrule=0.4pt, arc=4pt,
  left=8pt, right=8pt, top=6pt, bottom=6pt,
  before skip=8pt, after skip=8pt,
}

\NewDocumentEnvironment{findingsfigure}{+b}{%
  \afterpage{\afterpage{\begin{figure}[t]#1\end{figure}}}}{}
\newcommand{\coltablebox}[1]{\adjustbox{max width=\linewidth}{#1}}
\newlength{\colfigwidth}
\AtBeginDocument{\setlength{\colfigwidth}{0.465\textwidth}}

\title{\parbox{\linewidth}{\centering\setstretch{1}\fontsize{19}{25}\selectfont
  Harness Updating Is Not Harness Benefit:\\ Disentangling Evolution Capabilities\\ in Self-Evolving LLM Agents}\vspace{.5em}}

\author[1,*]{Minhua~Lin}
\author[2,*]{Juncheng~Wu}
\author[2]{Zijun~Wang}
\author[3]{Zhan~Shi}
\author[3]{Yisi~Sang}
\author[3]{Bing~He}
\author[4]{Zewen~Liu}
\author[5]{Tianxin~Wei}
\author[1]{Zongyu~Wu}
\author[1]{Zhiwei~Zhang}
\author[6]{Dakuo~Wang}
\author[1]{Xiang~Zhang}
\author[3]{Benoit~Dumoulin}
\author[2]{Cihang~Xie}
\author[2]{Yuyin~Zhou}
\author[1]{Suhang~Wang}
\author[3]{Hanqing~Lu}

\affiliation[1]{The Pennsylvania State University}
\affiliation[2]{UC Santa Cruz}
\affiliation[3]{Amazon}
\affiliation[4]{Emory University}
\affiliation[5]{UIUC}
\affiliation[6]{Northeastern University}
\contribution[*]{Equal contribution}

\abstract{\renewenvironment{abstract}{}{}\begin{abstract}
LLM agents are increasingly deployed as systems built around editable external harnesses, including prompts, skills, memories and tools, that shape task execution without changing model parameters. 
Harness self-evolution adapts such agents by updating these harnesses from execution evidence. 
Yet it remains unclear whether a model's \emph{base capability }in task-solving predicts its capabilities in harness self-evolution: which models produce useful harness updates, and which actually benefit from them?
We analyze two harness self-evolution capabilities: (i) \emph{harness-updating}, the capability to produce useful persistent harness updates from execution evidence; (ii) \emph{harness-benefit}, the capability to benefit from updated harnesses during task solving.
Our analysis reveals two findings.
First, \emph{harness-updating is flat in base capability}: models from different capability tiers produce harness updates that lead to surprisingly similar gains; even Qwen3.5-9B's updates yield gains comparable to those of Claude Opus~4.6.
Second, \emph{harness-benefit is non-monotonic in base capability}: weak-tier models benefit little from updated harnesses, mid-tier models benefit most, and strong-tier models benefit less than mid-tier.
We trace low gains at the weak tier to two failure modes: weak-tier models may fail to activate relevant harness artifacts, or activate them but fail to follow them faithfully.
These findings suggest investing capability budget in the task-solving agent rather than the evolver, and targeting harness invocation and long-horizon instruction following in agent training.
\codeavailability{Our source code is publicly available at \href{https://github.com/A-EVO-Lab/a-evolve/tree/release/harness-evolution}{here}.}

\end{abstract}
}
\correspondence{\email{luhanqin@amazon.com}}
\metadata[Project Page]{\url{https://ventr1c.github.io/HarnessDisentangle/}}
\newcommand{\codeavailability}[1]{}
\metadata[Code]{\url{https://github.com/A-EVO-Lab/a-evolve/tree/release/harness-evolution}}

\begin{document}
\maketitle

% The abstract is already in the title box (\abstract above).

\section{Introduction}
% \minhua{Background: What is agent harness and why is it important: LLMs power agentic systems; agentic systems depend on harnesses.}

% Large language models (LLMs)~\cite{radford2018improving,brown2020language,touvron2023llama} have become a general-purpose foundation for language understanding~\cite{wang2024mmlu,hendrycks2020measuring}, reasoning~\cite{wang2025evaluation}, and task solving~\cite{zhou2025engibench,lin2025wildbench}. Increasingly, they also power \emph{agentic systems} that interact with external environments, call tools, operate software interfaces, and complete long-horizon tasks~\cite{yao2022react,liu2024agentbench,yang2024swe,merrill2026terminal}. In these settings, system behavior depends not only on the underlying model but also on an external \emph{agent harness}: prompts~\cite{wei2022chain}, skills~\cite{xia2026skillrl}, memories~\cite{yan2025memory}, tools~\cite{qin2024toolllm},
% etc., that shape how the model observes, reasons, acts, and recovers from errors. Improving an agentic system increasingly means refining not only the foundation model, but also the editable harness around it.

Large language models (LLMs)~\cite{radford2018improving,touvron2023llama} have become a general-purpose foundation for language understanding~\cite{hendrycks2020measuring}, reasoning~\cite{wang2025evaluation}, and task solving~\cite{zhou2025engibench}. Increasingly, they also power \emph{agentic systems} that interact with external environments, call tools, operate software interfaces, and complete long-horizon tasks~\cite{yang2024swe,merrill2026terminal}. In these settings, system behavior depends not only on the underlying model but also on an external \emph{agent harness}: prompts~\cite{wei2022chain}, skills~\cite{xia2026skillrl}, memories~\cite{yan2025memory}, tools~\cite{qin2024toolllm}, etc., that shape how the model observes, reasons, acts, and recovers from errors. Improving an agentic system increasingly means refining not only the foundation model, but also the editable harness around it.

% [12]: narrow lines in amazonpaper.tex, where the figure is deferred to the "Despite this rapid progress"
% paragraph on page 2; wrapfig's own count leaves one empty narrow line under the caption.
% Revisit if the figure or the text before it changes.
\begin{frameworkfigure}[12]
    \centering
    \includegraphics[width=0.85\linewidth]{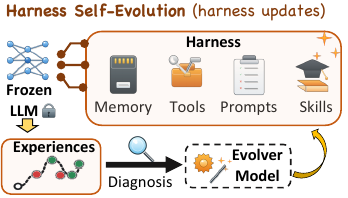}
    \caption{\textbf{Overview of harness self-evolution.}}
    \label{fig:overview_of_harness_evolution}
\end{frameworkfigure}

% findingsfigure: figure* in acl_latex.tex; amazonpaper.tex holds it back to the top of page 3, after Fig. 1.
\begin{findingsfigure}
\centering
\includegraphics[width=0.95\textwidth]{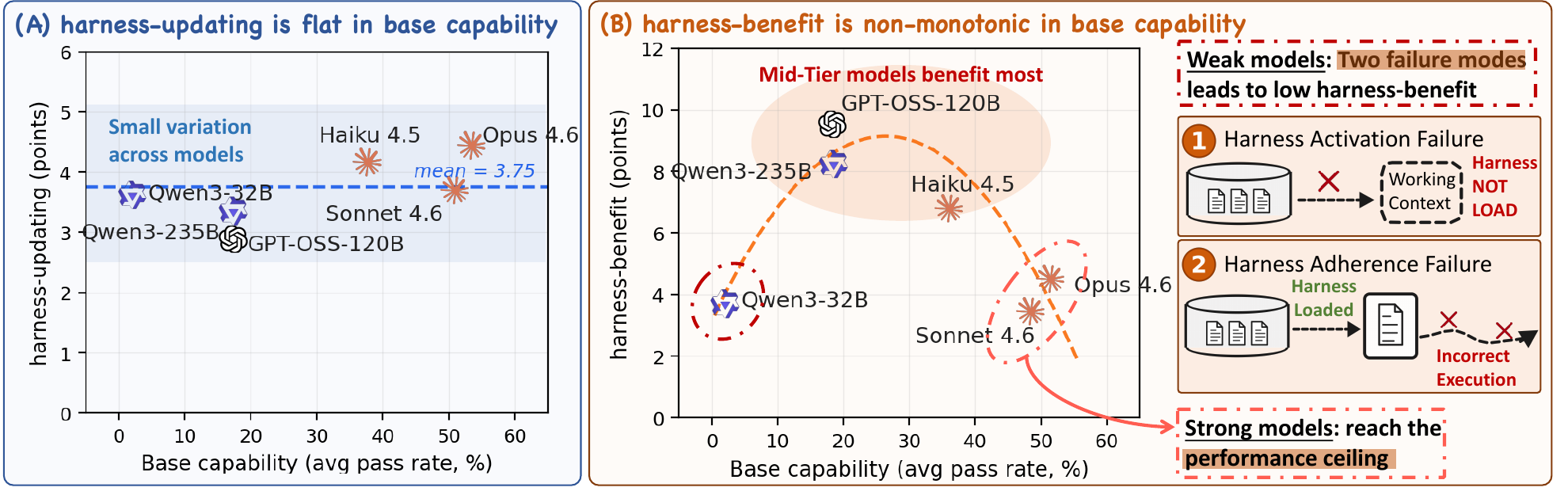}
\captionof{figure}{\textbf{Overview of our findings.}
\textbf{(i)} \emph{Harness-updating is flat in base capability}. Models across capability tiers produce harness updates that yield similar gains.
\textbf{(ii)} \emph{Harness-benefit is non-monotonic in base capability}. Mid-tier models benefit most, while weak-tier models benefit little due to failures in harness activation and adherence. 
% \suhang{It is unclear what the points means for the y-axis. Add gain??}\minhua{points mean the percentage points, harness-benifit and harness-updating already represent the performance gain by their the definition.}
}
\label{fig:findings}
\end{findingsfigure}

% \minhua{Background: Why do we need self evolution}
In current practice, harnesses are typically designed by hand. However, such manual design is brittle in deployment-time environments: task distributions shift, edge cases appear, and useful procedures are discovered only after the system interacts with real tasks.
A natural response is to update the harness automatically from execution evidence: failures, feedback, trajectories, and successful procedures can be written back into the harness and reused on future tasks. We refer to this setting as \emph{harness evolution} (Fig.~\ref{fig:overview_of_harness_evolution}): the model weights remain fixed, while the external agent harness is revised over time. Recent self-evolving agent methods~\cite{madaan2023self,wu2025evolver,agrawal2026gepa,xia2026skillrl,lin2026position} pursue this approach across diverse harness components and have shown end-task improvements over non-evolving baselines. 
In these works, harness updates are typically produced by an LLM from execution evidence; we refer to this update role as the \emph{evolver}.

% \minhua{Limitations of existing evaluation: end-to-end gains hide capability sources}

Despite this rapid progress, evaluation of these methods still asks an end-to-end question: does a self-evolution method effectively improve agent performance? 
This question is important, but it hides the source of improvement. 
The gain may come from the \emph{evolver} producing higher-quality harness updates, or from the task-solving agent using the updated harnesses more effectively during task solving. 
End-to-end scores cannot disentangle these contributions, leaving two practical questions open: \emph{which models produce useful harness updates, and which models benefit most from them?}
% \zhan {whats the meaning of "from their interaction"; And "The gain may come from the evolver producing higher-quality harness updates, or from the task-solving agent more effectively leveraging the updated harness during task solving." sounds better -- "leveraging" is more precise}

% Despite this rapid progress, evaluation of these methods still asks an end-to-end question: \emph{does a self-evolution method effectively improve agent performance?} This question is important, but it hides the source of improvement. The gain may come from the model that produces high-quality harness updates, or from the model that uses those updates well. End-to-end scores cannot disentangle these contributions, leaving two practical questions open: which models produce useful harness updates, and which models benefit most from them?

To answer these questions, we analyze \textbf{two evolution capabilities} a model exercises in harness self-evolution across three agentic benchmarks and seven LLMs: \emph{harness-updating}, the capability to produce useful harness updates from execution evidence; and \emph{harness-benefit}, the capability to benefit from updated harnesses during task solving. 
A model exercises harness-updating as the evolver, and harness-benefit as the task-solving agent.
% Harness-updating is exercised when a model acts as the evolver, and harness-benefit when it acts as the task-solving agent.
We conduct comprehensive experiments by pairing seven LLMs, spanning open-source and closed-source families across capability tiers, as agents and evolvers on three representative agentic benchmarks. Our analysis reveals two systematic decouplings between harness-evolution capabilities and \emph{base capability}, namely, a model's task-solving capability without harness evolution (Fig.~\ref{fig:findings}).

% Specifically, we conduct comprehensive experiments by pairing seven LLMs, spanning open-source and closed-source families across multiple capability tiers, as agents and evolvers on three representative agentic substrates: SWE-bench Verified~\cite{jimenez2024swe}, MCP-Atlas~\cite{bandi2026mcp}, and SkillsBench~\cite{li2026skillsbench}. Our analysis reveals two systematic decouplings between harness-evolution capabilities and \minhua{base task-solving ability} (Fig.~\ref{fig:findings}).

First, \textbf{harness-updating is flat in base capability}. When we fix the task-solving agent and vary the evolver model, models from different capability tiers produce harness updates that lead to surprisingly similar gains, and no evolver dominates across all substrates. Our case studies further show that even the Qwen3.5-9B evolver produces harness updates whose downstream gains match those of Claude Opus~4.6, despite a large gap in base capability.

Second, \textbf{harness-benefit is non-monotonic across base-capability tiers}.
% \textbf{harness-use is the stronger bottleneck, and the corresponding gain $\Delta_{\text{use}}$ is non-monotonic in base capability}:
Mid-tier models (e.g., GPT-OSS-120B) benefit most from updated harness, and strong-tier models (e.g., Claude Opus~4.6) reach the performance ceiling and benefit less. The weak-tier end, however, is not explained by the same ceiling argument: with the largest headroom above their base capability, models like Qwen3-32B might be expected to benefit most, yet they benefit the least. Our in-depth analysis identifies two failure modes that explain this weak-tier 
gap: (i) \emph{harness activation failure}: weak models often \emph{fail to invoke} relevant harness artifacts (e.g., skills) during task-solving; and (ii) \emph{harness adherence failure}: even when the harness is loaded, weak models \emph{fail to adhere} to it due to weak instruction-following over long-horizon tasks.

% Second, \textbf{harness-benefit is non-monotonic in base capability}.
% % \textbf{harness-use is the stronger bottleneck, and the corresponding gain $\Delta_{\text{use}}$ is non-monotonic in base capability}:
% Models with mid-tier base-capability (e.g., GPT-OSS-120B) benefit most from updated harness, and strong-tier models (e.g., Claude Opus~4.6) reach the performance ceiling and benefit less. The weak-tier end, however, is not explained by the same ceiling argument: with the largest headroom above their base capability, models like Qwen3-32B might be expected to benefit most, yet they benefit the least. To understand why weak-tier models gain little, our in-depth analysis identifies two failure modes: (i) \emph{harness activation failure}: weak models often \emph{fail to invoke} relevant harness artifacts (e.g., skills) during task-solving; and (ii) \emph{harness adherence failure}: even when the harness is loaded, weak models \emph{fail to adhere} to it due to weak instruction-following over long-horizon tasks.

% often reverting to their parametric knowledge due to weak instruction-following over long-horizon tasks.
% Together, these findings argue for evaluating harness evolution as two distinct capabilities of foundation models, separable from base capability, and motivate future work that targets each capability independently in model training and deployment.

These findings translate into design guidance for harness self-evolution systems. \emph{(i) Allocate capability budget to the task-solving agent, not the evolver}: 
the harness-updating gap across evolvers is at most 3.1~percentage points on 
any benchmark, so scaling up the evolver yields limited returns; post-evolution 
performance varies much more with the task-solving agent than with the evolver. \emph{(ii) Bake harness invocation into agent training}: weak-tier models often fail to load the harness at all (e.g., 25\% load rate for Qwen3-32B against $\approx96\%$ for strong models), so harness invocation should be treated as a first-class learned skill. \emph{(iii) Strengthen long-horizon instruction following}: even when loaded, weak-tier adherence decays across the trajectory over four times more steeply than strong models, making sustained instruction following a second key target for downstream agent training.

\section{Related Work}
\label{sec:related_works}
% \noindent\textbf{Harness engineering.}
% An LLM agent combines a frozen backbone with an external \emph{harness} that mediates reasoning, tool use, and environment interaction~\cite{yao2022react,yang2024swe,ning2026code}. Recent work treats the harness as a first-class design object, and the works differ primarily in how the artifacts the agent reads are represented. \textit{Prompts and instructions} remain the most direct form: \citet{pan2026natural} formalise the entire harness as natural-language artifacts that the model itself reads and writes. \textit{Skills} package reusable executable procedures into a callable library, of which \citet{wang2023voyager} is an early instance. \textit{Memory} stores across-task experience that can be retrieved into context, of which \citet{ouyang2025reasoningbank} is a representative instance that curates compact strategy-level items. \textit{Code} treats the harness itself as an executable program, of which \citet{lee2026meta} is a recent instance that searches over harness source with an agentic proposer.

\noindent\textbf{Harness engineering.}
An LLM agent combines a frozen backbone with an external \emph{harness} that mediates reasoning, tool use, memory access, and environment interaction~\cite{yao2022react,yang2024swe,ning2026code}. 
Recent work treats the harness as a first-class design object, differing mainly in the type of artifact exposed to the agent. 
\textit{Prompts and instructions} provide natural-language guidance~\cite{zhou2022large,pan2026natural}; 
\textit{tools} expose external services and define how agents discover, invoke, and validate them~\cite{hou2025model,qin2024toolllm,liu2025toolace,lin2026how}; 
\textit{memory} stores prior observations, facts, and strategies for later retrieval~\cite{ouyang2025reasoningbank,xu2026mem,fang2026lightmem}; 
\textit{skills} package reusable procedures into callable modules~\cite{li2026skillsbench,liu2026graph}; 
and \textit{code} treats the harness itself as executable source that can be optimized by an agentic proposer~\cite{lee2026meta}. 
These works establish harnesses as editable agent state. 
Our work shifts the focus from harness representation to model capabilities in updating and benefiting from harnesses. 
More details are in Appendix~\ref{app:related_works_harness}.

\noindent\textbf{Self-evolution of LLM agents.}
Beyond \emph{what} the harness contains, a complementary line asks how it is \emph{updated} from execution experience.
Early systems adapt agents through episode- or task-level language feedback: verbal self-reflection~\cite{shinn2023reflexion} and iterative self-feedback~\cite{madaan2023self} improve later attempts by feeding lessons back into context.
More recent methods make persistent harness components the unit of self-evolution, updating prompts~\cite{agarwal2024promptwizard,zhang2025agentic,agrawal2026gepa}, memories~\cite{wu2025evolver,zhang2025memevolve,lin2026memma}, skills~\cite{xia2026skillrl,alzubi2026evoskill,yang2026autoskill}, or tools~\cite{chen2025learning,li2026yunjue} from execution traces.
Collectively, these methods show that writing execution experience back into the harness can improve downstream task performance.
However, evaluations in this line typically report the end-to-end gain of one update procedure paired with one target agent on one substrate~\cite{li2026skillsbench,jiang2026sea,wei2025evo}.
Such scores conflate three sources of improvement: the agent's base capability, the evolver's \emph{harness-updating}, and the agent's \emph{harness-benefit}.
Our work complements these methods with a controlled analysis that varies task-solving agents and evolvers independently, measures harness-updating and harness-benefit separately, and tests whether either tracks base capability. More details in Appendix~\ref{app:related_works_evolution}.

\section{Harness-Evolution Capabilities}
\label{sec:preliminary_and_method}

To explore the evolution capabilities in harness self-evolution, we consider harness self-evolution, which adapts an LLM agent by updating the external harness around a fixed model during task execution: the agent attempts a stream of tasks and the harness is updated based on the agent's execution evidence. In this section, we formalize the harness-evolution protocol and define two evolution capabilities: \emph{harness-updating}, the ability to produce useful harness updates, and \emph{harness-benefit}, the ability to benefit from updated harnesses.

\subsection{Preliminaries: Harness State and Evolver}
\label{subsec:harness}

\noindent\textbf{Agent Harness.}
We use \emph{agent harness} to denote the external, non-parametric context and infrastructure through which an LLM is deployed for task execution~\cite{yao2022react,ning2026code,lee2026meta}. Formally, at evolution step
$t$, the LLM agent is defined as:
\begin{equation}
A_t = (f, H_t),
\end{equation}
where $f$ is the agent's model backbone and $H_t$ is the harness state after step $t$. Following common harness self-evolution settings~\cite{zhou2026externalization,lin2026position}, we keep $f$ fixed and only update editable components of $H_t$ (e.g., prompts, skills, memories), and fix other components such as tool interfaces and execution policies.

% \noindent\textbf{Evolver.}
% An \emph{evolver} is the update procedure that converts the agent's execution evidence into harness updates, where recent self-evolving agent systems~\cite{yang2024large,yuksekgonul2024textgrad,xia2026skillrl,agrawal2026gepa} increasingly instantiate this procedure with LLM agents. Formally, given the previous harness $H_{t-1}$ and the accumulated execution evidence $\mathcal{D}_t$ at step $t$, the evolver $e$ proposes a harness update and applies it to $H_{t-1}$ to obtain the next harness:
% \begin{equation}
% \label{eq:evolver_def}
% \begin{aligned}
% \Delta H_t &= e(H_{t-1}, \mathcal{D}_t; \beta),\\
% H_t &= \mathrm{Apply}(H_{t-1}, \Delta H_t).
% \end{aligned}
% \end{equation}
% where $\beta$ is the update budget \suhang{explain what the budget means in harness update???} and $\mathrm{Apply}$ denotes the commit operation to apply $\Delta H_t$ to $H_{t-1}$.

\noindent\textbf{Evolver.}
An \emph{evolver} is the update procedure that converts the agent's execution evidence into harness updates, where recent self-evolving agent systems~\cite{yang2024large,yuksekgonul2024textgrad,xia2026skillrl,agrawal2026gepa} increasingly instantiate this procedure with LLM agents. Formally, given the previous harness $H_{t-1}$ and the accumulated execution evidence $\mathcal{D}_t$ at step $t$, the evolver $e$ proposes a harness update and applies it to $H_{t-1}$ to obtain the next harness:
\begin{equation}
\label{eq:evolver_def}
\begin{aligned}
\Delta H_t &= e(H_{t-1}, \mathcal{D}_t),\\
H_t &= \mathrm{Apply}(H_{t-1}, \Delta H_t).
\end{aligned}
\end{equation}
where $\mathrm{Apply}$ denotes the commit operation to apply $\Delta H_t$ to $H_{t-1}$.

\subsection{Evolution Protocol}
\label{subsec:protocol}
Following common harness self-evolution pipelines~\cite{ouyang2025reasoningbank, agrawal2026gepa}, we formalize the protocol as an iterative loop between task-solving and harness evolution. Starting from an initial harness $H_0$, the protocol iterates for $T$ steps. At each step, the agent runs on a batch of tasks, collects execution evidence, and the evolver updates the harness for the next step.
Formally, given an agent $A_{t-1} = (f, H_{t-1})$ and a task batch $\mathcal{X}_t$ at step $t$, $A_{t-1}$ attempts to solve each task $x \in \mathcal{X}_t$ and output:
\begin{equation}
(\tau_{t,x}, y_{t,x}) = \mathrm{Solve}(A_{t-1}, x)
\end{equation}
where $\tau_{t,x}$ is the execution trajectory and $y_{t,x}$ is the final output. The execution evidence $\mathcal{D}_t$ is then:
\begin{equation}
\mathcal{D}_t = \{(x, \tau_{t,x}, y_{t,x}) : x \in \mathcal{X}_t\}.
\end{equation}
The evolver produces the updated harness $H_t$ from $H_{t-1}$ and $\mathcal{D}_t$ as in Eq.~\ref{eq:evolver_def}, yielding the next agent $A_t = (f, H_t)$. This loop repeats for $T$ steps, producing the final harness $H_T$.
\subsection{Capability Metrics}
\label{subsec:capability}
To analyze which models produce useful harness updates and which models benefit from them, we formally define three metrics to measure both harness-evolution capabilities (i.e., \emph{harness-updating} and \emph{harness-benefit}) along with each model's \emph{base capability}.

\noindent\textbf{Base Capability and Evolution Gain.}
Given a task set $\mathcal{X} = \bigcup_{t=1}^{T} \mathcal{X}_t$, the \emph{base capability} of a model $f$ is the task-solving performance of the initial agent $A_0 = (f, H_0)$ on $\mathcal{X}$:
\begin{equation}
M_{\text{base}}(f) = J_{\mathcal{X}}(f, H_0),
\end{equation}
where $J_{\mathcal{X}}(f, H)$ is the scoring function that measures the performance of agent $(f, H)$ on $\mathcal{X}$. 

Given a model $f$ and an evolver $e$, let $H_T^{(f, e)}$ denote the final harness produced after evolution with $f$ as the agent and $e$ as the evolver for $T$ steps starting from $H_0$. We further define the \emph{pairwise evolution gain} as the improvement of a specific agent--evolver pairing $(f, e)$ over the agent's task-solving performance before evolution:
\begin{equation}
\label{eq:pairwise_evolution_gain}
\Delta(f, e) = J_{\mathcal{X}}(f, H_T^{(f, e)}) - M_{\text{base}}(f).
\end{equation}

\noindent\textbf{Harness-updating Capability.}
The \emph{harness-updating capability} of an evolver $e$ is its ability to produce harness updates that improve agents' task-solving. Formally, this is defined as the mean pairwise gain across an anchor agent set $\mathcal{F}^\star$:
\begin{equation}
\Delta_{\text{update}}(e) = \frac{1}{|\mathcal{F}^{\star}|} \sum_{f \in \mathcal{F}^{\star}} \Delta(f, e).
\end{equation}

\noindent\textbf{Harness-benefit Capability.}
The \emph{harness-benefit capability} of a model $f$ is its maximum gain in task-solving performance from harness self-evolution. In practice, we estimate this as the maximum pairwise gain across a fixed anchor evolver set $\mathcal{E}^\star$:
\begin{equation}
\Delta_{\text{benefit}}(f) = \max_{e \in \mathcal{E}^{\star}} \Delta(f, e).
\end{equation}

\section{Experiments}
\label{sec:experiments}

In this section, we empirically analyze the two harness-evolution capabilities defined in Sec.~\ref{sec:preliminary_and_method}. We present the evolver-side analysis of harness-updating capability in Sec.~\ref{subsec:efficacy_results}, and the agent-side analysis of harness-benefit capability in Sec.~\ref{subsec:harness_using_results}

\subsection{Experimental Setup}
\label{subsec:setup}

\noindent\textbf{Datasets.} We evaluate on three representative agentic benchmarks: SWE-bench Verified (SWE)~\cite{jimenez2024swe} for software engineering, MCP-Atlas (MCP)~\cite{bandi2026mcp} for tool use over real MCP servers, and SkillsBench (SB)~\cite{li2026skillsbench} for skill-based execution across diverse domains. More details of these datasets are in Appendix~\ref{app:datasets}.

\begin{figure*}[t]
\centering
\includegraphics[width=\linewidth]{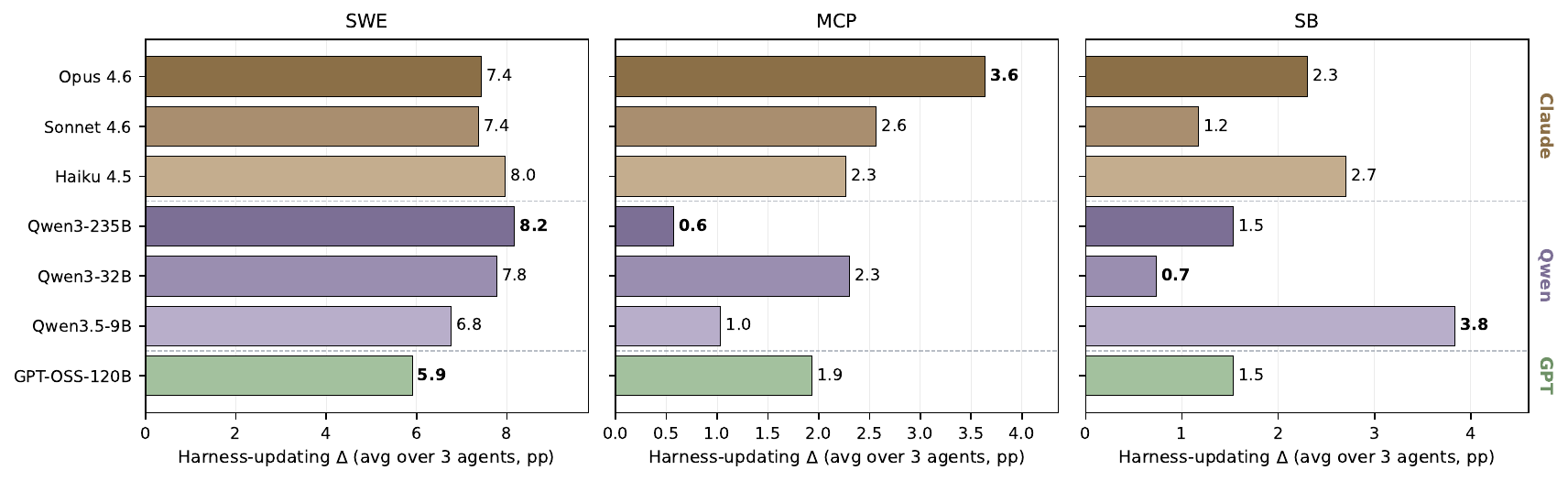}
\caption{\textbf{Harness-updating capability ($\Delta_{\text{update}}$) of each evolver.} Evolvers are grouped by model family (Claude, Qwen, GPT-OSS). The best and worst evolver, marked in bold within each panel, change with the benchmark.}
\label{fig:evolver_delta_bars}
\end{figure*}

% \noindent\textbf{Models.}
% We use seven LLMs as backbones for both task-solving agents and evolvers, spanning open and closed source families across capability tiers: Claude Opus~4.6~\cite{anthropic2026claudeopus46}, Claude Sonnet~4.6~\cite{anthropic2026claudesonnet46}, Claude Haiku~4.5~\cite{anthropic2025claudehaiku45}, Qwen3-235B-A22B and Qwen3-32B~\cite{yang2025qwen3}, Qwen3.5-9B~\cite{qwen35blog}, and GPT-OSS-120B~\cite{agarwal2025gpt}.
% More details are in Appendix~\ref{app:models}.

\noindent\textbf{Models.}
We use seven LLM backbones, spanning open-source and closed-source families across capability tiers.
For the agent-side analysis, we use six models: Claude Opus~4.6~\cite{anthropic2026claudeopus46}, Claude Sonnet~4.6~\cite{anthropic2026claudesonnet46}, Claude Haiku~4.5~\cite{anthropic2025claudehaiku45}, Qwen3-235B-A22B and Qwen3-32B~\cite{yang2025qwen3}, and GPT-OSS-120B~\cite{agarwal2025gpt}.
For the evolver-side analysis, we use the same six models plus Qwen3.5-9B~\cite{qwen35blog}, the smallest model in this paper, to test whether a substantially smaller open model can still produce useful harness updates.

% MiniMax-M2.5, and Kimi-K2.5~\cite{team2026kimi}.  

\noindent\textbf{Evaluation Protocol.}
We report three metrics defined in Sec.~\ref{subsec:capability}: base capability $M_{\mathrm{base}}(f)$, harness-updating gain $\Delta_{\mathrm{update}}(e)$, and harness-benefit gain $\Delta_{\mathrm{benefit}}(f)$. To calculate them, we use pass rate as the primary metric for $J_{\mathcal{X}}$ on three benchmarks. We consider an in-situ evaluation setting: each task in $\mathcal{X}_t$ is scored under $H_{t-1}$ before its evidence is used to produce $H_t$. The final results are reported by aggregating per-task scores over the task stream. Further details are in Appendix~\ref{app:metrics}.

\noindent\textbf{Implementation Details.}
We instantiate the evolution protocol in Sec.~\ref{subsec:protocol} with a fixed solve-evolve loop. 
For a fair comparison, 
we fix the prompt template for both agents and evolvers, along with the trajectory window, 
across all agent-evolver pairs; only the LLM backbone varies. All pairs within a benchmark 
start from the same initial harness $H_0$ and task stream $\mathcal{X}$, share the same 
evolution budget $\beta$ and per-task turn limit. The evolvable components are skills 
for SWE-bench Verified and SkillsBench, and skills, prompts, and memories for MCP-Atlas. 
Other details such as prompt templates are in Appendix~\ref{app:implementation_details}.

\subsection{Evolver-side Analysis}
\label{subsec:efficacy_results}
To understand how harness-updating capability varies across LLMs, we fix the task-solving agents and vary the evolver over the seven LLMs in Sec.~\ref{subsec:setup}. 
Specifically, we use three representative LLMs, Opus~4.6, Sonnet~4.6, and Qwen3-235B, as the anchor agents in $\mathcal{F}^{\star}$. 
For each evolver $e$, we report $\Delta_{\mathrm{update}}(e)$, defined in Sec.~\ref{subsec:capability}, across the three benchmarks in Fig.~\ref{fig:evolver_delta_bars}. Full pass-rate results for all agent-evolver pairings are in Appendix~\ref{app:evolver_delta_update}.

\noindent\textbf{Observation~1: Harness-updating is flat in base capability.}
Fig.~\ref{fig:evolver_delta_bars} shows two patterns:
\textbf{(i)} \emph{The spread of $\Delta_{\text{update}}$ across evolvers is narrow}. The gap between the best and worst evolver is at most 3.1~percentage points (pp) on any benchmark, and no model wins across benchmarks. Qwen3-235B illustrates this reshuffling: it leads on SWE (8.2~pp) but ranks last on MCP (0.6~pp). 
\textbf{(ii)} \emph{Model scale is not predictive}. The smallest evolver, Qwen3.5-9B, posts the highest gain on SB (3.8~pp), exceeding both Opus~4.6 (2.3~pp) and Qwen3-235B (1.5~pp). 

\begin{figure*}[t]
\centering
\includegraphics[width=\linewidth]{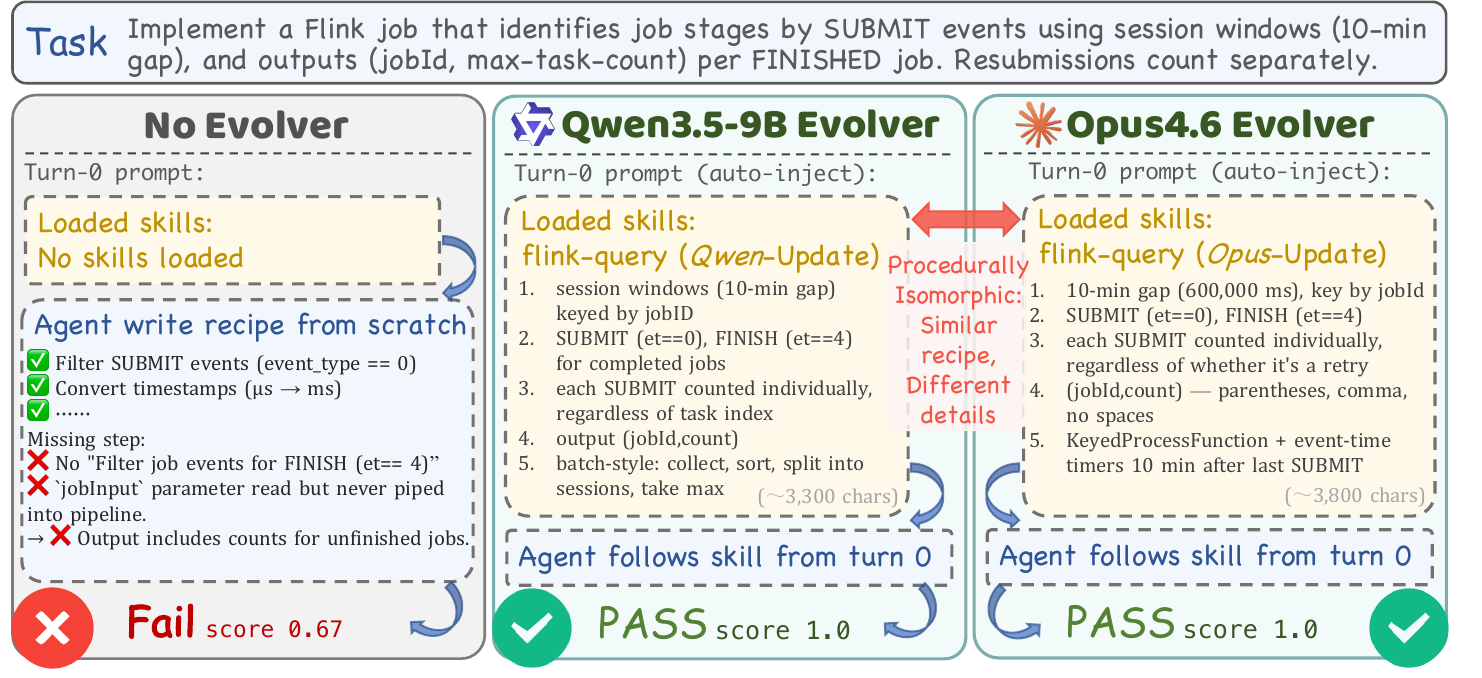}
\caption{\textbf{Comparison of harness updated by Qwen3.5-9B and Claude Opus 4.6.} 
We compare an Opus~4.6 agent on the SkillsBench \texttt{flink-query} task under three conditions: no evolved skill (left, score 0.67), a skill evolved by Qwen3.5-9B (center, score 1.0), and a skill evolved by Opus~4.6 (right, score 1.0). 
Both evolved skills encode procedurally similar guidance and enable the same agent to solve the task.
% Trajectory of an agent solving the SkillsBench \texttt{flink-query} task under three turn-0 conditions: no evolver (left, score 0.67), with a skill evolved by Qwen3.5-9B injected (center, score 1.0), and with a skill evolved by Opus~4.6 injected (right, score 1.0). 
}
\label{fig:evolver_flink_case}
\end{figure*}

\noindent\textbf{Case Study: the 9B evolver writes a skill procedurally isomorphic to Opus's.}
To understand the mechanism behind these comparable gains, we examine a representative SkillsBench task \texttt{flink-query} in detail. We fix the task-solving agent backbone at Opus~4.6 and compare its trajectories under three evolver conditions (Fig.~\ref{fig:evolver_flink_case}): no evolver, Qwen3.5-9B as evolver, and Opus~4.6 as evolver. We observe that without an evolved skill, the agent omits the FINISH-event filter and fail to solve this task (scores 0.67); with a skill injected by either Qwen3.5-9B or Opus~4.6, the same agent solves the task successfully (score 1.0). Inspecting the two skills, we find they are procedurally isomorphic, prescribing the same sequence of steps and differing only in surface details of implementation and verbosity. The 9B open-source evolver thus reaches the same procedural content as the frontier evolver.
Full details of the skill contents and analysis are in Appendix~\ref{app:more_details_case_study_evolver_side}.

\begin{figure}[t]
\centering
\includegraphics[width=0.9\colfigwidth]{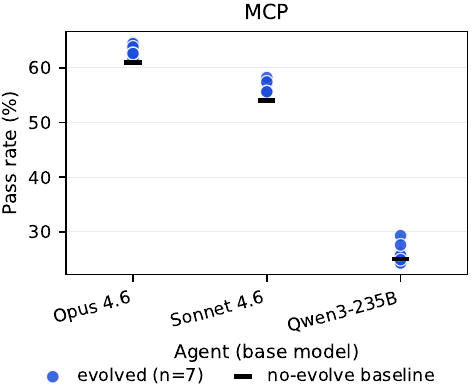}
% \vspace{-0.5em}
\caption{\textbf{MCP post-evolution scores}: for each anchor agent every blue dot is one of seven evolved scores and the black tick is the no-evolve baseline. Within-agent variation across evolvers is small relative to between-agent variation in base capability.}
\label{fig:evolver_agent_dominance_mcp}
\end{figure}

\noindent\textbf{Observation~2: Post-evolution score is dominated by models' base capability, not evolver identity.}
To understand the relative contribution of task-solving agents and evolvers to post-evolution performance, we plot the task-solving performances of three LLMs (Opus~4.6, Sonnet~4.6, Qwen3-235B) in $\mathcal{F}^{\star}$ under the updated harnesses from seven LLMs in Sec.~\ref{subsec:setup} as the evolvers against each agents' base capability. Results on MCP-Atlas are shown in Fig.~\ref{fig:evolver_agent_dominance_mcp}. We observe: 
\textbf{(i)} \emph{Within-agent spread is much smaller than between-agent gap.} The within-agent spread across seven evolvers is at most 5.1~pp (Qwen3-235B), small against the 36.0~pp gap between the Opus and Qwen3-235B base capabilities. The pattern persists on SWE and SB.
\textbf{(ii)} \emph{Extreme pairing still favors strong agents.} Even pairing the weakest anchor agent with its best-performing evolver against the strongest anchor agent with its worst-performing evolver, the strong agent still leads by 18.6 to 35.2~pp on every benchmark. 
Both patterns also persist on SWE and SB datasets (Appendix~\ref{app:more_analysis_evolver_side_finding2}).
Post-evolution performance is therefore bottlenecked on the agent side, not the evolver side, motivating the agent-side analysis in Sec.~\ref{subsec:harness_using_results}.

\noindent\textbf{Take-away.} Allocate capability budget to the task-solving agent, not the evolver: (i) $\Delta_{\text{update}}$ varies by at most 3.1~pp across evolvers on any benchmark, and (ii) post-evolution score is dominated by the agent's base capability.

% \takeaway{Allocate capability budget to the task-solving agent, not the evolver: (i) $\Delta_{\text{update}}$ varies by at most 3.1~pp across evolvers on any benchmark, and (ii) post-evolution score is dominated by the agent's base capability.}

\subsection{Agent-side Analysis}
\label{subsec:harness_using_results}
\begin{figure}[t]
\centering
\includegraphics[width=0.85\colfigwidth]{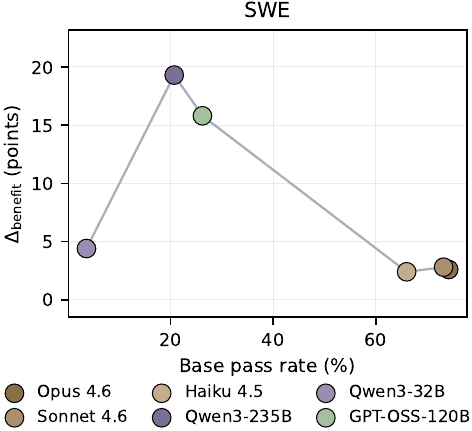}
% \vspace{-0.5em}
% \caption{\textbf{$\Delta_{\mathrm{benefit}}$ versus base pass rate on SWE for six task-solving agents.} 
% Each point corresponds to one LLM backbone. 
% Points are connected in ascending order of base pass rate. 
% MCP and SB analogues are in Appendix~\ref{app:agent_dh_extra}.}
\caption{\textbf{$\Delta_{\mathrm{benefit}}$ versus base pass rate on SWE.} Each point is one LLM backbone used as the task-solving agent; points are connected in ascending base pass rate. MCP and SB analogues are in Appendix~\ref{app:agent_dh_extra}.}
\label{fig:agent_dh_swe}
\end{figure}

To understand how \emph{harness-benefit} capability varies across LLMs, we fix the evolvers and vary the task-solving agent over the LLM backbones in Sec.~\ref{subsec:setup}: Opus~4.6, Sonnet~4.6, Haiku~4.5, Qwen3-235B, Qwen3-32B, and GPT-OSS-120B. 
We use Opus~4.6, Sonnet~4.6, and Qwen3-235B as the three anchor evolvers, denoted by $\mathcal{E}^{\star}$. 
For each agent $f$, we report $\Delta_{\mathrm{benefit}}(f)$, defined in Sec.~\ref{subsec:capability}, in Tab.~\ref{tab:agent_delta_harness} 
and Fig.~\ref{fig:agent_dh_swe}. The full pass-rate results for all agent-evolver pairings are in Tab.~\ref{tab:full_agent_evolver_matrix} in Appendix~\ref{app:agent_dh_extra}.

\noindent\textbf{Observation~1: $\Delta_{\mathrm{benefit}}$ is non-monotonic in base capability.}
As shown in Tab.~\ref{tab:agent_delta_harness} and Fig.~\ref{fig:agent_dh_swe}, $\Delta_{\mathrm{benefit}}$ does not increase monotonically with base capability. 
On SWE, the gain peaks at Qwen3-235B (19.3~pp), while the weaker Qwen3-32B gains only 4.4~pp and the stronger Opus~4.6 gains only 2.6~pp. 
On MCP, the peak shifts to GPT-OSS-120B (7.0~pp), again with lower gains at both ends of the base-capability scale. 
This pattern has different explanations at the two ends of the capability scale. 
At the high-capability end, smaller gains are consistent with a ceiling effect: strong models already solve many tasks under the initial harness, leaving less room for further improvement. However, at the low-capability end, smaller gains reflect a different bottleneck, which we diagnose next.

\begin{table}[t]
\centering
\small
\setlength{\tabcolsep}{4pt}
% \caption{\textbf{Base pass rate (\%) and Harness-benefit $\Delta_{\text{benefit}}$ (pp) per agent per benchmark.} Agents are sorted by SWE base ascending. Bold marks the column maximum of $\Delta_{\text{benefit}}$.}

% \caption{\textbf{Base pass rate (\%) and harness-benefit $\Delta_{\mathrm{benefit}}$ (pp) for six agents across benchmarks.} 
% Each row is one LLM backbone used as the task-solving agent. 
% Models are sorted by SWE base pass rate. 
% Bold marks the largest $\Delta_{\mathrm{benefit}}$ value within each benchmark.}
\caption{\textbf{Base pass rate (\%) and harness-benefit $\Delta_{\mathrm{benefit}}$ (pp) across benchmarks.} Each row is one LLM backbone used as the task-solving agent. Bold marks the largest $\Delta_{\mathrm{benefit}}$ within each benchmark.}

\begin{tabular}{l|rr rr rr}
\toprule
 & \multicolumn{2}{c}{\textbf{SWE}} & \multicolumn{2}{c}{\textbf{MCP}} & \multicolumn{2}{c}{\textbf{SB}} \\
\cmidrule(lr){2-3} \cmidrule(lr){4-5} \cmidrule(lr){6-7}
\textbf{Model} & Base & $\Delta$ & Base & $\Delta$ & Base & $\Delta$ \\
\midrule
Qwen3-32B    &  3.6 &  4.4 &  3.6 &  1.0 &  0.0 &  5.8 \\
Qwen3-235B   & 20.7 & \textbf{19.3} & 25.0 &  4.3 &  4.7 &  1.1 \\
GPT-OSS-120B & 26.2 & 15.8 & 28.0 & \textbf{7.0} &  0.0 &  7.0 \\
Haiku 4.5    & 66.0 &  2.4 & 42.4 &  3.6 &  5.8 & \textbf{15.1} \\
Sonnet 4.6   & 73.2 &  2.8 & 54.0 &  3.2 & 24.4 &  3.5 \\
Opus 4.6     & 74.2 &  2.6 & 61.0 &  3.6 & 25.6 &  5.8 \\
\bottomrule
\end{tabular}
\label{tab:agent_delta_harness}
\end{table}

\begin{figure*}[t]
\centering
\includegraphics[width=1\linewidth]{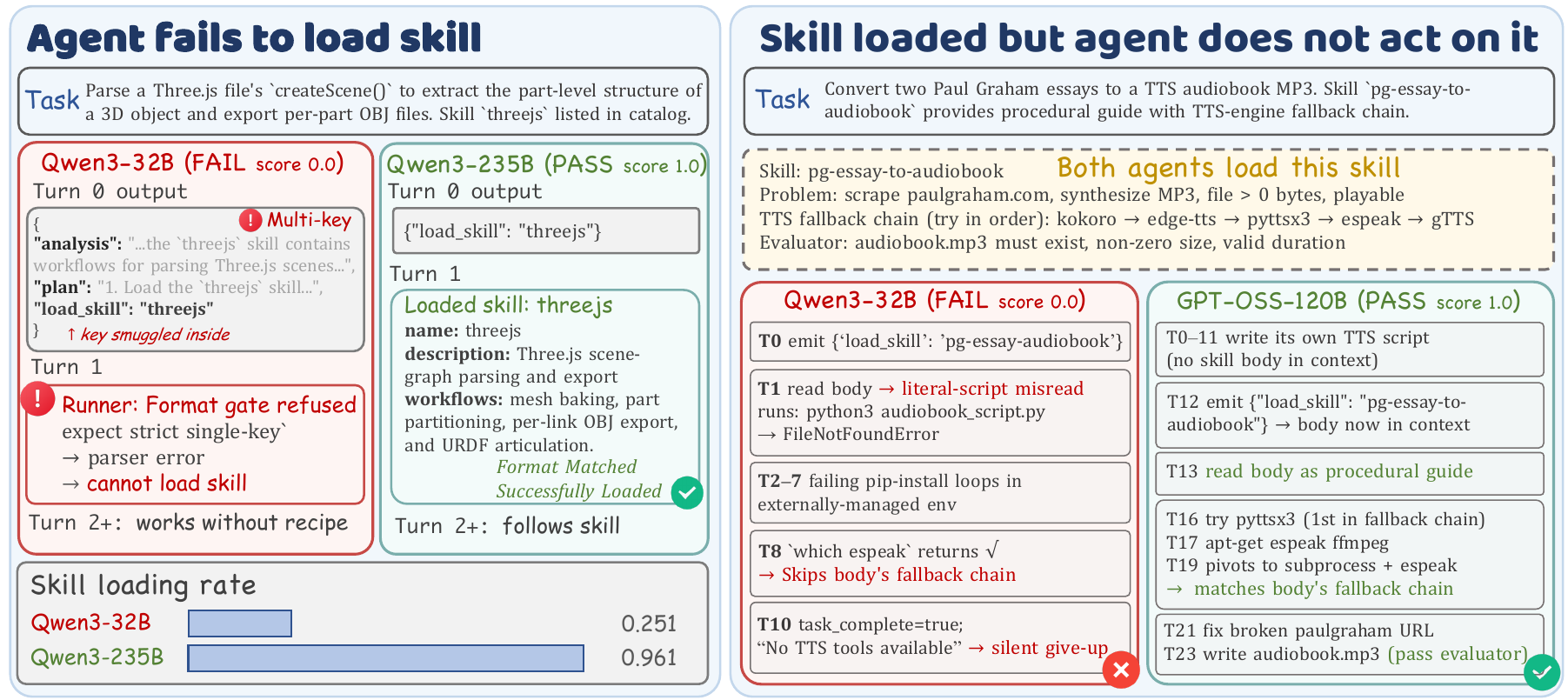}
\caption{\textbf{Two harness-benefit failure modes for Qwen3-32B on SkillsBench.} 
Left (\texttt{threejs}): \emph{harness activation failure}, where an invalid multi-key load action prevents the skill body from entering context. 
Right (\texttt{pg-essay-to-audiobook}): \emph{harness adherence failure}, where the skill is loaded but the agent treats it as a literal script and skips the prescribed fallback chain.}
\label{fig:solver_side_failure}
\end{figure*}

\noindent\textbf{Observation~2: Weak-tier models derive low $\Delta_{\text{benefit}}$ due to two failure modes.}
To understand why the weak-tier models  with low base capabilities receive low $\Delta_{\text{benefit}}$, we conduct an in-depth analysis on SkillsBench and identify two complementary failure modes: \emph{harness activation} and \emph{harness adherence}, which is illustrated in Fig.~\ref{fig:solver_side_failure}. 

The first mode is \emph{harness activation failure}: weak-tier models often fail to bring relevant harness artifacts, such as skills, into their working context.
To quantify this on SkillsBench, we report each agent's \emph{skill-load rate (SLR)}, the fraction of its trajectories in which it actively loads at least one skill into its context.
Tab.~\ref{tab:agent_sfr} shows that the skill-load rate is near ceiling for Opus~4.6, Sonnet~4.6, and Qwen3-235B (0.957--0.961), but drops to 0.446 for GPT-OSS-120B and 0.251 for Qwen3-32B.
The left panel of Fig.~\ref{fig:solver_side_failure} illustrates this activation failure. Specifically, Qwen3-32B identifies the relevant skill, but embeds the loading request inside a broader action rather than issuing it as a standalone skill-loading action.
The SkillsBench environment therefore does not treat it as a valid load request, so the skill body never enters context.

The second mode is \emph{harness adherence failure}: even when relevant harness artifacts are loaded, weak-tier models often fail to follow their guidance faithfully during task solving. 
We quantify this failure with the \emph{Harness-Following Rate} (HFR), computed over trajectories in which at least one skill is loaded. 
For each skill-loaded task-solving trajectory, an LLM judge determines whether the task-solving model follows the loaded skill's guidance. HFR is the fraction of skill-loaded trajectories judged as following the skill. Appendix~\ref{app:hfr_pipeline} provides details of the judge pipeline.
Tab.~\ref{tab:agent_sfr} reports HFR together with two complementary metrics: \emph{SLR}, which measures harness activation, and \emph{pass-when-loaded (LPR)}, which measures the pass rate among that model's skill-loaded trajectories. 
We observe two patterns. 
\textbf{(i)} \emph{Strong-tier models exhibit much higher harness adherence than weak-tier models.} 
Opus~4.6 reaches an HFR of 0.757, while Qwen3-32B reaches only 0.142. 
\textbf{(ii)} \emph{Loading the harness is not sufficient for benefiting from it.} 
Qwen3-235B provides the cleanest separation between activation and adherence: its skill-load rate is 0.961, nearly identical to Opus~4.6, yet its HFR is only 0.350. 
Its pass-when-loaded rate mirrors this gap, at 0.022 compared with 0.177 for Opus~4.6.
The \texttt{pg-essay-to-audiobook} case in the right panel of Fig.~\ref{fig:solver_side_failure} illustrates this adherence failure. 
Qwen3-32B successfully loads the procedural skill, but treats the guidance as a ready-made script rather than a procedure to follow. 
After the first attempt fails, it terminates instead of trying the alternative steps prescribed by the skill.
More details of the analysis are in Appendix~\ref{app:more_details_case_study_agent_side}.

\begin{table}[t]
\centering
\small
\setlength{\tabcolsep}{5pt}
\caption{\textbf{Per-model activation, adherence, and outcome metrics on SkillsBench.}
\textit{SLR}: fraction of a model's trajectories in which at least one skill is loaded into context.
\textit{HFR}: fraction of skill-loaded trajectories judged as following the loaded skill's guidance.
\textit{LPR}: pass rate among the model's skill-loaded trajectories. 
Models are sorted by base capability on SkillsBench.}
\begin{tabular}{l|ccc}
\toprule
\textbf{Model} & \textbf{SLR} & \textbf{HFR} & \textbf{LPR} \\
\midrule
Qwen3-32B    & \textit{0.251} & \textit{0.142} & 0.023 \\
GPT-OSS-120B & 0.446 & 0.442 & 0.040 \\
Haiku 4.5    & 0.794 & 0.600 & 0.099 \\
Qwen3-235B   & \textbf{0.961} & 0.350 & 0.022 \\
Sonnet 4.6   & 0.959 & 0.730 & 0.145 \\
Opus 4.6     & 0.957 & \textbf{0.757} & \textbf{0.177} \\
\bottomrule
\end{tabular}
\label{tab:agent_sfr}
\end{table}

\noindent\textbf{Diagnosis: Weak instruction following over long-horizon execution.}
To test whether harness adherence degrades as a trajectory unfolds, we conduct a phase-level adherence analysis. 
An LLM judge assigns a 0--1 adherence score at different execution stages, with details provided in Appendix~\ref{app:judge_details_phase_adherence_score}. 
We use Qwen3-32B, GPT-OSS-120B, and Opus~4.6 as representative weak-, mid-, and strong-tier models, respectively. 
Tab.~\ref{tab:phase_drift_representative} reports three representative phases, after harness loading, at the trajectory midpoint, and at final validation, with scores averaged over judged trajectories for each model. We observe that
Qwen3-32B drops sharply from 0.52 after harness loading to 0.13 at final validation, while GPT-OSS-120B drops more moderately from 0.67 to 0.43. 
In contrast, Opus~4.6 remains stable, from 0.89 to 0.80. 
This graded drift suggests a long-horizon instruction-following bottleneck: weaker models progressively lose adherence as the trajectory unfolds, rather than merely misreading the harness at load time.

\begin{table}[t]
\centering
\small
\setlength{\tabcolsep}{3.5pt}
% \caption{\textbf{Per-phase adherence score for representative weak-, mid-, and strong-tier models.}
% Qwen3-32B, GPT-OSS-120B, and Claude Opus~4.6 are used as representative weak-, mid-, and strong-tier models. Bold and underline indicate the best and the worst scores among models in each phase.}
\caption{\textbf{Per-phase adherence scores for representative weak-, mid-, and strong-tier models}
(Qwen3-32B, GPT-OSS-120B, and Opus~4.6). 
\textbf{Bold} and \underline{underlining} mark the best and worst score in each phase.}
\begin{tabular}{l|ccc}
\toprule
\textbf{Trajectory Phase}            & \textbf{Qwen3-32B}   & \textbf{GPT-OSS} & \textbf{Opus 4.6} \\
                            & (weak)      & (mid)        & (strong) \\
\midrule
Harness loaded    & \underline{0.52}        & 0.67         & \textbf{0.89} \\
Mid turn                    & \underline{0.22}        & 0.48         & \textbf{0.79} \\
Final turn       & \underline{0.13}        & 0.43         & \textbf{0.80} \\
\midrule
drift (load $\to$ final)    & \underline{-0.39}     & -0.24      & \textbf{-0.09} \\
\bottomrule
\end{tabular}
\label{tab:phase_drift_representative}
\end{table}

% \noindent\textbf{Take-away.} Agent training should target harness-benefit along two axes:
% (i) \emph{Bake harness invocation into training.} Targets low harness-load rate at the weak end (0.25 for Qwen3-32B against $\approx 0.96$ for strong agents);
% (ii) \emph{Strengthen long-horizon instruction following.} Targets the trajectory-tail drift (weak-agent adherence drifts $-0.34$ from load to final, three times more than strong agents).

\noindent\textbf{Take-away.} Agent training should target harness-benefit along two axes.
(i) \emph{Bake harness invocation into training}: weak-tier models have low skill-load rates (25.1\% for Qwen3-32B vs.\ $\approx96\%$ for strong-tier models), so agents must learn to reliably bring relevant harness artifacts into context. 
\emph{(ii) Strengthen long-horizon instruction following}: even after loading the harness, weak-tier models lose adherence over the trajectory (Qwen3-32B drifts from 0.52 to 0.13), so agents must learn to sustain harness guidance over long-horizon tasks.

\section{Conclusion}
We analyze harness self-evolution by decomposing it into two model capabilities distinct from base capability: \emph{harness-updating}, the capability to produce harness updates, and \emph{harness-benefit}, the capability to benefit from updated harnesses during task solving. 
Across seven LLMs and three benchmarks, harness-updating is flat in base capability: models across capability tiers produce updates that yield similar gains, and even the Qwen3.5-9B evolver induces gains comparable to Claude Opus~4.6. 
In contrast, harness-benefit is non-monotonic in base capability: weak-tier models gain little, traced to two failure modes: failing to activate relevant harness artifacts and failing to follow them faithfully once activated. 
% In contrast, harness-benefit is non-monotonic in base capability: weak-tier models gain little, traced to two failure modes, failing to activate relevant harness artifacts or failing to follow them faithfully once activated. 
These findings motivate investing capability budget in the agent rather than the evolver, and targeting agent training at harness invocation and long-horizon instruction following.

% We analyze harness self-evolution of LLM agents by decomposing it into two model capabilities distinct from base capability: \emph{harness-updating}, the capability to produce useful harness updates from execution evidence, and \emph{harness-benefit}, the capability to benefit from updated harnesses during task solving. Across seven LLMs on three benchmarks, harness-updating is flat in base capability: models from different capability tiers produce harness updates that yield surprisingly similar gains, and a 9B open model matches frontier proprietary models in update quality. In contrast, harness-benefit is non-monotonic in base capability: weak-tier models often gain little from evolved harnesses, which we trace to two failure modes: weak-tier models may fail to activate relevant harness artifacts, or activate them but fail to follow them faithfully. These findings translate into design guidance for harness self-evolution systems: capability budget should be invested in the agent rather than the evolver, and agent training should target two axes, harness invocation and long-horizon instruction following, to close the weak-tier gap.

\section{Limitations}

Our study focuses on harness self-evolution, where model weights remain fixed and adaptation occurs through updates to external harness artifacts. 
We do not evaluate parametric fine-tuning, reinforcement learning of model weights, or hybrid adaptation methods that combine weight updates with harness updates. 
Our model set is representative but not exhaustive: we include open-source and closed-source models across multiple capability tiers, but a broader model grid would further clarify how harness-updating and harness-benefit vary with model family, scale, training recipe, and deployment cost.

\section{Ethics Statement}

This work studies LLM agents that update persistent external harnesses from execution evidence. 
All experiments are conducted on benchmark tasks, and we do not collect or process private user data. 
However, harness self-evolution raises broader deployment concerns because updated harnesses may persist across future tasks. 
Incorrect lessons, unsafe tool-use rules, biased instructions, or sensitive information could be written into the harness and reused by later agents.
In our evaluation, harness updates are logged, and evolvers are constrained from modifying evaluation scripts or updating model weights. 
These controls make the benchmark setting auditable, but they do not by themselves guarantee safety in open deployments. 
Real-world harness self-evolution systems should treat privacy, consent for data retention, update reversibility, auditability, and human oversight as first-class design requirements.

\clearpage
\bibliographystyle{assets/plainnat}
\bibliography{acl_latex}

\clearpage
\beginappendix
% \section{Evolution Algorithm Details}
% \label{app:evolution_framework_details}

\section{Full Details of Related Works}
\label{app:additional_related_work}

In this section, we provide the full version of the related works in Sec.~\ref{sec:related_works}.

\subsection{Harness Engineering}
\label{app:related_works_harness}

LLM agents are increasingly deployed as compound systems in which a frozen model is surrounded by external artifacts that shape reasoning, tool use, memory access, skill invocation, and environment interaction.
We refer to this external layer as the agent harness.
Prior work studies several forms of harness artifacts.
\textit{Prompts} encode standing behavioral rules, task policies, and reasoning procedures in natural language~\cite{zhou2022large,yao2022react,yang2024swe,pan2026natural}.
\textit{Tools} expose external services and specify the action schemas, invocation formats, and validation rules through which agents interact with them~\cite{hou2025model,qin2024toolllm,liu2025toolace,lin2025comprehensive,lin2026how,weng2026s,lin2026decoding}.
\textit{Memory} stores prior observations, facts, task outcomes, and reusable strategies for later retrieval or consolidation~\cite{ouyang2025reasoningbank,xu2026mem,fang2026lightmem}.
\textit{Skills} package reusable procedures into callable modules or task-specific guidance artifacts, as studied in skill benchmarks and skill-library systems~\cite{li2026skillsbench,liu2026graph}.
\textit{Code} treats the harness itself as executable source that can implement tools, validators, orchestration logic, and prompt assembly~\cite{ning2026code,lee2026meta}.

These works establish harnesses as editable agent state rather than passive context.
Our work is complementary: instead of proposing a new harness representation, we analyze the model capabilities involved in updating harness artifacts and benefiting from the resulting updates.

\subsection{Self Evolution of LLM agents}
\label{app:related_works_evolution}
Beyond \emph{what} the harness contains, a complementary line asks how harness artifacts are updated from execution experience. 
Early systems operate at the task-attempt level. 
Reflexion~\cite{shinn2023reflexion} stores verbal self-reflections from prior attempts, Self-Refine~\cite{madaan2023self} iteratively improves outputs through self-feedback, and ExpeL~\cite{zhao2024expel} extracts reusable natural-language insights from training trajectories for later retrieval. 
These methods show that language feedback can improve future behavior, but the persistent artifact is usually a single textual reflection or lesson, rather than a structured, multi-component harness state.

More recent methods make persistent harness components the unit of self-evolution~\cite{weng2026group,liu2026adaptive,wei2026evo}. 
\textit{Prompt-level} methods update natural-language instructions or prompt programs: PromptWizard~\cite{agarwal2024promptwizard} refines prompts through feedback-driven critique and synthesis, ACE~\cite{zhang2025agentic} evolves contextual playbooks through structured generation, reflection, and curation, and GEPA~\cite{agrawal2026gepa} evolves prompts through trajectory-level reflection. 
\textit{Memory-level} methods write experience into persistent stores that can be retrieved, refined, or reorganized across future tasks: EvolveR~\cite{wu2025evolver} connects offline strategy distillation with online retrieval, MemEvolve~\cite{zhang2025memevolve} studies meta-evolution of agent memory systems, and MemMA~\cite{lin2026memma} improves long-horizon memory through construction, retrieval, and feedback-driven repair. 
\textit{Skill- and workflow-level} methods package successful behavior into reusable procedures: Voyager~\cite{wang2023voyager} accumulates executable skills, AWM~\cite{wang2024agent} induces workflows from successful trajectories, SkillRL~\cite{xia2026skillrl} recursively expands a skill library through reinforcement learning, and EvoSkill~\cite{alzubi2026evoskill} studies automated skill discovery from agent experience. 
\textit{Tool-level} self-evolution further allows agents to synthesize, revise, or accumulate tools and tool-use knowledge over time~\cite{chen2025learning,li2026yunjue}. 

Collectively, these methods show that writing execution experience back into persistent harness components can improve downstream task performance. 
However, their evaluations typically report the end-to-end gain of one update procedure paired with one agent on one benchmark~\cite{li2026skillsbench,jiang2026sea,wei2025evo}. 
Such scores often conflate multiple sources of improvement: the agent's base capability under the initial harness, the evolver's \emph{harness-updating} capability in producing useful harness updates, and the agent's \emph{harness-benefit} capability in acting on those updates. 
Our work complements this line with a controlled capability analysis: we vary agents and evolvers independently, measure harness-updating and harness-benefit separately, and test whether either capability simply tracks base capability.

\begin{table*}[t]
\centering
\small
\caption{\textbf{Dataset statistics.} $N_b$ is the number of tasks; the rightmost column lists the static resources each task exposes to the agent.}
\label{tab:dataset_details}
\begin{tabular}{lccp{0.36\linewidth}}
\toprule
\textbf{Substrate} & $N_b$ & \textbf{\#Domains} & \textbf{Resources per task} \\
\midrule
SWE-bench Verified & $500$ & $12$ repositories & Codebase snapshot, issue description, hidden test suite \\
MCP-Atlas          & $500$ & $36$ MCP servers  & $220$ tools (shared across servers); $3$--$6$ tool calls required per task \\
SkillsBench        & $86$  & $11$ task domains & Workspace files, deterministic verifier \\
\bottomrule
\end{tabular}
\end{table*}

\section{Experimental Setup Details}
\subsection{Dataset Details}
\label{app:datasets}

We evaluate on three representative agentic benchmarks that cover complementary agent capabilities: long-horizon code repair with SWE-bench Verified, multi-server tool orchestration with MCP-Atlas, and skill-based execution across diverse domains with SkillsBench. 
Dataset statistics are in Tab.~\ref{tab:dataset_details}:
\begin{itemize}[leftmargin=*]
    \item \textbf{SWE-bench Verified~\cite{jimenez2024swe}.}
This is a human-validated subset of SWE-bench containing $500$ tasks drawn from real GitHub issues across $12$ popular Python repositories. Each task provides a codebase snapshot and an issue description; the solver must produce a patch that resolves the issue. A task passes if its patch satisfies the hidden test suite associated with the issue. We use the full $500$-task subset.
\item \textbf{MCP-Atlas~\cite{bandi2026mcp}.}
This is a benchmark for multi-server tool-use competency over real Model Context Protocol servers. Each task is a natural-language request whose completion requires the solver to identify and orchestrate $3$--$6$ tool calls across $36$ real MCP servers exposing $220$ tools. Scoring uses a claims-based rubric that awards credit per factual claim satisfied in the final answer; we report pass rate as the fraction of tasks for which all claims are satisfied. We use the $500$-task public subset released by the authors.
\item \textbf{SkillsBench~\cite{li2026skillsbench}.}
This is a $86$-task benchmark spanning $11$ domains (e.g., software, data analysis, document processing, audio synthesis) with a deterministic per-task verifier. Each task provides workspace files and a natural-language instruction; the agent must complete the task using the workspace and any skills available in its harness. The native benchmark ships with curated skills, but in our setup the no-evolution baseline starts from an empty skill set, and evolved cells use only the skills produced by the evolver from earlier in-situ tasks.
\end{itemize}

\subsection{Models}
\label{app:models}
We use seven LLM backbones, spanning open-source and closed-source families across capability tiers. The closed-source models are Claude Opus~4.6~\cite{anthropic2026claudeopus46}, Claude Sonnet~4.6~\cite{anthropic2026claudesonnet46}, and Claude Haiku~4.5~\cite{anthropic2025claudehaiku45}. The open-source models are Qwen3-235B-A22B and Qwen3-32B~\cite{yang2025qwen3}, Qwen3.5-9B~\cite{qwen35blog}, and GPT-OSS-120B~\cite{agarwal2025gpt}.

For the agent-side analysis, we use the six LLMs (Opus~4.6, Sonnet~4.6, Haiku~4.5, Qwen3-235B-A22B, Qwen3-32B, GPT-OSS-120B) as task-solving agent backbones. For the evolver-side analysis, we use all seven models, including Qwen3.5-9B (the smallest model in our paper), to test whether a substantially smaller open model can still produce useful harness updates. Across all experiments we query each model through its official API or inference endpoint; no model weights are updated during evolution.

\subsection{Metrics}
\label{app:metrics}

\noindent\textbf{Scoring function.}
For all four metrics in §\ref{subsec:capability}, we use pass rate as the scoring function $J_{\mathcal{X}}$: each task $x \in \mathcal{X}$ receives a per-task score from the benchmark's grader, and $J_{\mathcal{X}}$ is the mean over $\mathcal{X}$. Pass rates and average scores are reported in percent; gains are reported in percentage points.

\noindent\textbf{Per-benchmark scoring.}
The scoring function $J_{\mathcal{X}}$ instantiates the standard grading procedure of each benchmark:
\begin{itemize}[leftmargin=*]
\item \textbf{SWE-bench Verified}~\cite{jimenez2024swe}: per-task binary resolved score (1 if the submitted patch passes the designated fail-to-pass and pass-to-pass test suite, 0 otherwise). The mean over tasks is the standard pass rate.
\item \textbf{MCP-Atlas}~\cite{bandi2026mcp}: per-task claim-fulfillment score in $[0, 1]$, computed as the fraction of reference claims satisfied by the agent's final answer. We report both the strict pass rate (mean of binarized per-task scores) and the average claim-fulfillment score (mean of continuous per-task scores).
\item \textbf{SkillsBench}~\cite{li2026skillsbench}: per-task binary score averaged over $5$ trials following Terminal-Bench~\cite{merrill2026terminal}. We report the average score (mean across tasks and trials) as the primary metric.
\end{itemize}
For each benchmark, $J_{\mathcal{X}}$ in the metric definitions of §\ref{subsec:capability} refers to the mean of these per-task scores aggregated over the task stream.

\noindent\textbf{In-situ evaluation.}
We evaluate in an in-situ setting: the same task stream $\mathcal{X} = \bigcup_{t=1}^{T} \mathcal{X}_t$ that drives evolution also serves as the evaluation set. Concretely, at step $t$, each task $x \in \mathcal{X}_t$ is scored under the harness $H_{t-1}$ at the time of its attempt; the score is locked in before $(\tau_{t, x}, y_{t, x})$ enters $\mathcal{D}_t$ and produces $H_t$. The pass rate of any individual task is thus not influenced by harness updates derived from that task itself.

% \subsection{}

% \subsection{Implementation Details}
% \label{app:implementation_details}

% \minhua{Agent framework}
% \noindent\textbf{Prompt Template.}

\begin{table}[t]
\centering
\small
\setlength{\tabcolsep}{4pt}
% \caption{\textbf{Pass rate (\%) of each anchor agent under each evolver, across the three benchmarks.} The \textsc{None} row is the no-evolution baseline. $\Delta_{\text{update}}$ is the mean improvement of the three anchors over \textsc{None}, in percentage points. \textbf{Bold}: per-benchmark best evolver. \underline{Underline}: per-benchmark worst evolver.}
\caption{\textbf{Full evolver-side matrix.}
Within each benchmark block, entries under the three anchor agents are pass rates (\%) for that agent-evolver pairing; the $\Delta_{\text{update}}$ column reports the corresponding harness-updating score (pp); see Sec.~\ref{subsec:capability}. 
The \textsc{None} row is the no-evolution baseline. 
\textbf{Bold} and \underline{underlining} in the $\Delta_{\mathrm{update}}$ column mark the best and worst evolvers, respectively.}
\coltablebox{%
\begin{tabular}{l rrr r}
\toprule
Evolver & Opus~4.6 & Sonnet~4.6 & Qwen3-235B & $\Delta_{\text{update}}$ \\
\midrule
\multicolumn{5}{l}{\emph{SWE}} \\
\textsc{None} & 74.2 & 73.2 & 20.7 & --- \\
Opus~4.6      & 76.4 & 76.0 & 38.0 & 7.4 \\
Sonnet~4.6    & 76.8 & 75.6 & 37.8 & 7.4 \\
Haiku~4.5     & 77.8 & 74.8 & 39.4 & 8.0 \\
Qwen3-235B    & 76.6 & 76.0 & 40.0 & \textbf{8.2} \\
Qwen3-32B     & 76.2 & 75.4 & 39.8 & 7.8 \\
Qwen3.5-9B    & 76.4 & 73.2 & 38.8 & 6.8 \\
GPT-OSS-120B  & 75.2 & 75.6 & 35.0 & \underline{5.9} \\
\midrule
\multicolumn{5}{l}{\emph{MCP}} \\
\textsc{None} & 61.0 & 54.0 & 25.0 & --- \\
Opus~4.6      & 64.4 & 57.2 & 29.3 & \textbf{3.6} \\
Sonnet~4.6    & 64.6 & 57.0 & 26.1 & 2.6 \\
Haiku~4.5     & 64.4 & 58.2 & 24.2 & 2.3 \\
Qwen3-235B    & 61.6 & 55.8 & 24.3 & \underline{0.6} \\
Qwen3-32B     & 63.8 & 57.4 & 25.7 & 2.3 \\
Qwen3.5-9B    & 62.6 & 55.6 & 24.9 & 1.0 \\
GPT-OSS-120B  & 62.6 & 55.6 & 27.6 & 1.9 \\
\midrule
\multicolumn{5}{l}{\emph{SB}} \\
\textsc{None} & 25.6 & 24.4 &  4.7 & --- \\
Opus~4.6      & 30.2 & 27.9 &  3.5 & 2.3 \\
Sonnet~4.6    & 29.1 & 25.6 &  3.5 & 1.2 \\
Haiku~4.5     & 31.4 & 25.6 &  5.8 & 2.7 \\
Qwen3-235B    & 31.4 & 22.1 &  5.8 & 1.5 \\
Qwen3-32B     & 30.2 & 22.1 &  4.6 & \underline{0.7} \\
Qwen3.5-9B    & 26.7 & 31.4 &  8.1 & \textbf{3.8} \\
GPT-OSS-120B  & 31.4 & 22.1 &  5.8 & 1.5 \\
\bottomrule
\end{tabular}%
}
\label{tab:evolver_delta_update}
\end{table}

\subsection{Implementation Details}
\label{app:implementation_details}

\noindent\textbf{Evolvable Harness Artifacts}
The editable harness scope is benchmark-specific. 
SWE-bench Verified and SkillsBench allow edits only to the \texttt{skills} directory, while MCP-Atlas additionally allows edits to \texttt{prompts/system.md} and append-only updates to \texttt{memory/} JSONL files. 
The \texttt{tools/} directory and evaluation files are read-only for all benchmarks. 
These permissions are passed to the evolver at each cycle; the evolver system prompt itself is fixed across benchmarks and model backbones.

\noindent\textbf{Task-solving Agent Prompt Templates.}
Within each benchmark, all task-solving agents use the same system prompt; only the task-specific user prompt varies across tasks. 
For SWE-bench Verified, the solver prompt (Tab.~\ref{tab:prompt_solver_swe}) is an 828-byte procedural guide that scopes the agent to GitHub-issue patching and encourages minimal, focused edits. 
For MCP-Atlas, the solver prompt (Tab.~\ref{tab:prompt_solver_mcp}) is a 1{,}309-byte API-agent guide that instructs the agent to satisfy task queries through tool calls and not ask the user for clarification. 
For SkillsBench, we follow the original setting~\cite{li2026skillsbench} to use no system prompt for the task-solving agent.

\noindent\textbf{Evolver Prompt Template.}
All evolver backbones use the same system prompt, shown in Tab.~\ref{tab:prompt_evolver_default}. 
At each evolution cycle, the user message follows a fixed wrapper containing the cycle index, the writable-scope block, and the canonicalized execution-evidence payload. 
Thus, across benchmarks and model backbones, the prompt format is fixed; only the task evidence and benchmark-specific writable scope vary.

\section{Evolver-side Analysis Details in Sec.~\ref{subsec:efficacy_results}}
\label{app:evolver_side_analysis_details}

\subsection{Additional Results for Observation~1}
\label{app:evolver_delta_update}

Tab.~\ref{tab:evolver_delta_update} reports the pass rate of each anchor agent (Opus~4.6, Sonnet~4.6, Qwen3-235B) under each evolver on the three benchmarks, alongside the resulting $\Delta_{\text{update}}$. These are the per-cell numbers underlying the bars in Fig.~\ref{fig:evolver_delta_bars}.

\begin{figure*}[t]
\centering
\includegraphics[width=0.9\linewidth]{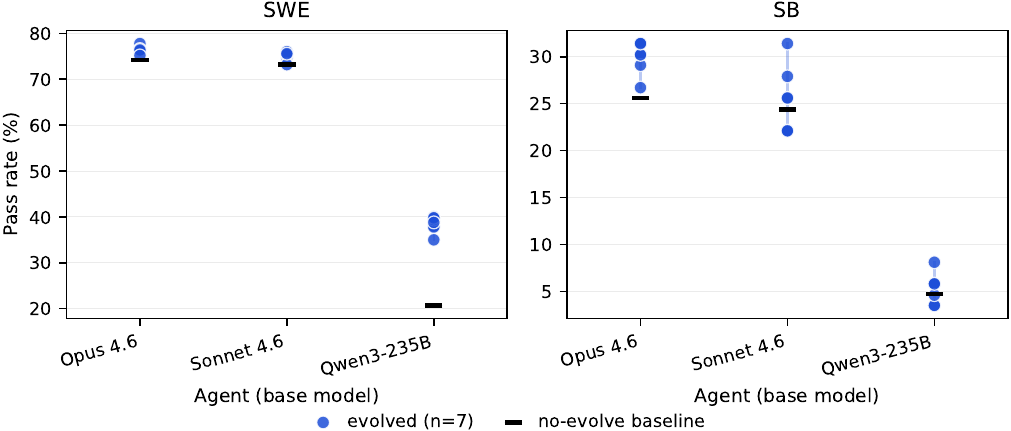}
\caption{\textbf{Post-evolution scores across evolvers for anchor agents on SWE (left) and SB (right) datasets.} 
Each anchor task-solving agent is instantiated with a different LLM backbone: Opus~4.6, Sonnet~4.6, or Qwen3-235B. 
Blue dots show scores obtained with the seven evolvers, and the black tick marks the no-evolution baseline.}
\label{fig:evolver_agent_dominance_appendix}
\end{figure*}

\begin{table*}[t]
\centering
\small
\setlength{\tabcolsep}{4pt}
\caption{\textbf{Extreme agent-evolver pairings across benchmarks.}
For each benchmark, $W$ is the weakest anchor task-solving agent and $S$ is the strongest anchor task-solving agent.
We pair $W$ with its best-performing evolver and $S$ with its worst-performing evolver among the seven evolvers.
Scores are pass rates (\%); the gap is the strong-agent score minus the weak-agent score, reported in percentage points (pp).}

\begin{tabular}{lccc}
\toprule
                                    & SWE  & MCP  & SB   \\
\midrule
weak anchor agent $W$                     & Q3-235B & Q3-235B & Q3-235B \\
best evolver for $W$                      & Q3-235B & Opus & Q3.5-9B \\
score of $W$ with best evolver            & 40.0 & 29.3 & 8.1  \\
\midrule
strong anchor agent $S$                   & Opus & Opus & Opus \\
worst evolver for $S$                     & GPT-OSS & Q3-235B & Q3.5-9B \\
score of $S$ with worst evolver           & 75.2 & 61.6 & 26.7 \\
\midrule
gap: strong-worst minus weak-best (pp)    & \textbf{35.2} & \textbf{32.3} & \textbf{18.6} \\
\bottomrule
\end{tabular}
\label{tab:agent_dominance}
\end{table*}

\subsection{More Details of the Case Study}
\label{app:more_details_case_study_evolver_side}

We elaborate on the case study from Sec.~\ref{subsec:efficacy_results}. We examine the SkillsBench task \texttt{flink-query} with the agent backbone fixed at Opus~4.6, comparing its trajectories under three evolver conditions (Fig.~\ref{fig:evolver_flink_case}): no evolver, Qwen3.5-9B as evolver, and Opus~4.6 as evolver. Without an evolver, the agent omits the FINISH-event filter and scores 0.67; with either evolved skill injected at turn~0, the same agent solves the task (score 1.0).

Inspecting the two evolved skills, we find that they encode the same five problem-solving steps:
\begin{itemize}[leftmargin=*]
    \item Filter SUBMIT events.
    \item Filter FINISH events.
    \item Count each SUBMIT separately.
    \item Emit \texttt{(jobId, count)}.
    \item Apply a 10-minute session window.
\end{itemize}
The two skills differ only in implementation surface details: Qwen3.5-9B specifies the gap as 10 minutes with manual batch sessionization, while Opus~4.6 specifies 10 minutes with a \texttt{KeyedProcessFunction}. Despite these surface differences, both skills yield identical downstream pass rates (1.0) when injected into the same Opus~4.6 agent.

\begin{table*}[t]
\centering
\small
% \caption{\textbf{Per-cell pass rate (\%) and harness-benefit $\Delta_{\text{benefit}}$ (pp).} Columns are task-solving agents, sorted ascending by SWE base. Rows are evolvers, with \textsc{None} the no-evolution baseline. $\Delta_{\text{benefit}}$ is the best evolver's gain over \textsc{None} for that agent and matches the column $\Delta$ values in Tab.~\ref{tab:agent_delta_harness}.}
\caption{\textbf{Full agent-side matrix underlying $\Delta_{\text{benefit}}$.}
Each cell reports pass rate (\%) for a task-solving model under a given evolver. 
The \textsc{None} row is the no-evolution baseline. 
$\Delta_{\text{benefit}}$ is the maximum gain over \textsc{None} across the three anchor evolvers, reported in percentage points (pp). 
\textbf{Bold} marks the largest $\Delta_{\text{benefit}}$ value in each benchmark block, and \underline{underlining} marks the smallest.}
\label{tab:full_agent_evolver_matrix}
\resizebox{\textwidth}{!}{
\begin{tabular}{ll cccccc}
\toprule
\textbf{Benchmark} & \textbf{Evolver} & \textbf{Qwen3-32B} & \textbf{Qwen3-235B} & \textbf{GPT-OSS-120B} & \textbf{Haiku 4.5} & \textbf{Sonnet 4.6} & \textbf{Opus 4.6} \\
\midrule
\multirow{5}{*}{SWE-bench Verified}
& \textsc{None}                       & 3.6  & 20.7 & 26.2 & 66.0 & 73.2 & 74.2 \\
& Opus 4.6                            & 8.0  & 38.0 & 37.2 & 65.0 & 76.0 & 76.4 \\
& Sonnet 4.6                          & 7.6  & 37.8 & 37.6 & 68.4 & 75.6 & 76.8 \\
& Qwen3-235B                          & 8.0  & 40.0 & 42.0 & 65.4 & 76.0 & 76.6 \\
& \textbf{$\Delta_{\text{benefit}}$}  & 4.4  & \textbf{19.3} & 15.8 & \underline{2.4}  & 2.8  & 2.6  \\
\midrule
\multirow{5}{*}{MCP-Atlas}
& \textsc{None}                       & 3.6  & 25.0 & 28.0 & 42.4 & 54.0 & 61.0 \\
& Opus 4.6                            & 4.6  & 29.3 & 35.0 & 46.0 & 57.2 & 64.4 \\
& Sonnet 4.6                          & 4.0  & 26.1 & 32.0 & 42.8 & 57.0 & 64.6 \\
& Qwen3-235B                          & 2.8  & 24.3 & 29.1 & 41.0 & 55.8 & 61.6 \\
& \textbf{$\Delta_{\text{benefit}}$}  & \underline{1.0}  & 4.3  & \textbf{7.0}  & 3.6  & 3.2  & 3.6  \\
\midrule
\multirow{5}{*}{SkillsBench}
& \textsc{None}                       & 0.0  & 4.7  & 0.0  & 5.8  & 24.4 & 25.6 \\
& Opus 4.6                            & 3.5  & 3.5  & 7.0  & 20.9 & 27.9 & 30.2 \\
& Sonnet 4.6                          & 3.5  & 3.5  & 4.6  & 18.6 & 25.6 & 29.1 \\
& Qwen3-235B                          & 5.8  & 5.8  & 7.0  & 15.1 & 22.1 & 31.4 \\
& \textbf{$\Delta_{\text{benefit}}$}  & 5.8  & \underline{1.1}  & 7.0  & \textbf{15.1} & 3.5  & 5.8  \\
\bottomrule
\end{tabular}}
\end{table*}

\begin{figure*}[t]
\centering
\includegraphics[width=\linewidth]{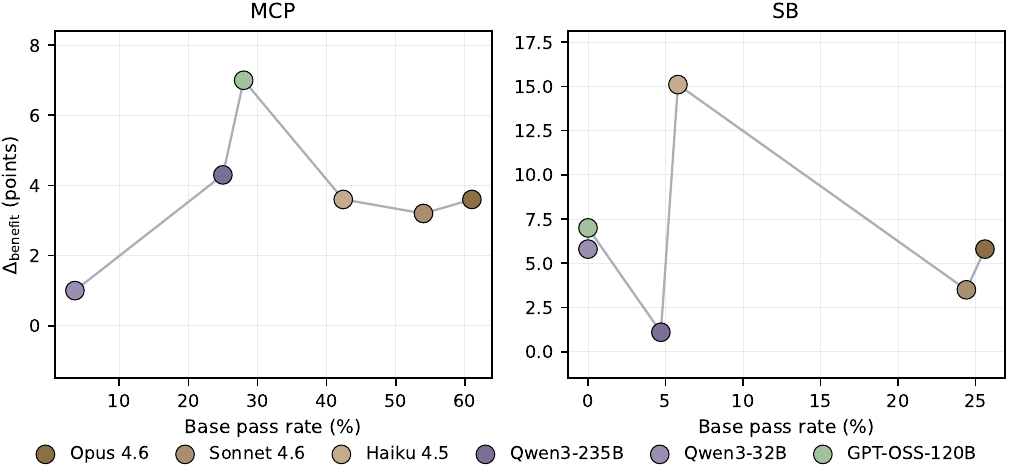}
\caption{\textbf{$\Delta_{\text{benefit}}$ versus base pass rate on MCP (left) and SB (right) datasets.} Each point corresponds to one LLM backbone used as the task-solving agent; points are connected in ascending base pass rate.}
\label{fig:agent_dh_appendix}
\end{figure*}

\subsection{Additional Results for Observation~2}
\label{app:more_analysis_evolver_side_finding2}
This subsection extends Observation~2 in Sec.~\ref{subsec:efficacy_results} to the other two benchmarks, SWE-bench Verified and SkillsBench. 
We observe the same two patterns: within-agent variation across evolvers remains smaller than between-agent differences in base capability, and even extreme agent-evolver pairings still favor the stronger agent.

\noindent\textbf{Within-agent spread versus between-agent gap.}
Fig.~\ref{fig:evolver_agent_dominance_appendix} extends the post-evolution score view of Fig.~\ref{fig:evolver_agent_dominance_mcp} to SWE and SB. 
On SWE, the largest within-agent spread across seven evolvers is 5.0~pp, attained by Qwen3-235B. 
On SB, the largest spread is 9.3~pp, attained by Sonnet~4.6, whose evolved scores range from 22.1\% to 31.4\%. 
By comparison, the base-capability gap between Opus~4.6 and Qwen3-235B is 53.5~pp on SWE and 20.9~pp on SB. 
Thus, the between-agent gap exceeds the within-agent spread by a factor of 11 on SWE and 2.2 on SB. 
SB is the tightest of the three benchmarks, but the same inequality still holds.

\noindent\textbf{Extreme pairings across benchmarks.}
Tab.~\ref{tab:agent_dominance} compares the weakest anchor agent $W$ paired with its best-performing evolver against the strongest anchor agent $S$ paired with its worst-performing evolver, separately for each benchmark. 
Even under this unfavorable comparison for the strong agent, $S$ still outperforms $W$ by 18.6 to 35.2~pp on every benchmark. 
On SB, the same evolver, Qwen3.5-9B, appears on both sides of the comparison, because it is the best evolver for Qwen3-235B and the worst evolver for Opus~4.6. 
This reinforces the main conclusion that post-evolution performance is dominated more by the task-solving agent than by evolver identity.

\section{Agent-side Analysis Details in Sec.~\ref{subsec:harness_using_results}}
\label{app:agent_side_analysis_details}

\subsection{Case Studies for the Two Agent-Side Failure Modes}
\label{app:more_details_case_study_agent_side}

We elaborate on the two failure cases in Fig.~\ref{fig:solver_side_failure}, both produced by Qwen3-32B on SkillsBench under the same harness and runner.

\noindent\textbf{Activation Failure: \texttt{threejs}.}
At turn~0, Qwen3-32B correctly identifies the relevant skill, but instead of emitting \texttt{load\_skill} as a standalone action, it produces a single multi-key JSON action that bundles \texttt{analysis} (free-form reasoning), \texttt{plan} (a step list), and \texttt{load\_skill}. The SkillsBench format gate accepts only single-key actions and rejects this composite as malformed. The skill body never enters the agent's context, and the agent proceeds without the procedural guidance the harness was meant to provide. The failure is at the action-protocol layer: the agent knows which skill to load, but cannot translate that intent into the runner's expected format.

\noindent\textbf{Adherence Failure: \texttt{pg-essay-to-audiobook}.}
The loaded skill prescribes a TTS-fallback chain: try a primary text-to-speech route, then fall back to alternative routes if the primary fails. Qwen3-32B successfully loads the skill at turn~0, but treats the chain as a literal script to execute rather than a contingent procedure. The first prescribed step hits a \texttt{FileNotFoundError} on turn~1; the agent then continues through subsequent turns without ever invoking the fallback steps. By turn~10, the agent emits \texttt{task\_complete:true} despite the absence of a valid task output, ending the trajectory below grader threshold. The failure is at the procedural-execution layer: the agent has loaded the skill but does not follow its contingent structure under unexpected runtime conditions.

\noindent\textbf{Common pattern.}
Both cases show that Qwen3-32B's weak-tier deficits are not in task understanding (it identifies the right skill in \texttt{threejs}; it follows the skill's first step in \texttt{pg-essay-to-audiobook}) but in protocol-level and procedural execution. This pattern is consistent with the activation and adherence trends in Tab.~\ref{tab:agent_sfr} and the per-phase drift in Tab.~\ref{tab:phase_drift_representative}: weak-tier models do not fail to read the harness, they fail to \emph{operate} under it.

% \section{Extended Result Tables}
\subsection{More results of $\Delta_{\text{benefit}}$ in Sec.~\ref{subsec:harness_using_results}}
\label{app:agent_dh_extra}
\noindent\textbf{Full Agent-Evolver Pass Rate.} 
Tab.~\ref{tab:full_agent_evolver_matrix} reports the full pass-rate matrix underlying the $\Delta_{\text{benefit}}$ values in Tab.~\ref{tab:agent_delta_harness}. 
For each benchmark and task-solving model, we report the no-evolution baseline and the pass rate under each of the three anchor evolvers, $\mathcal{E}^{\star}=\{\text{Opus~4.6}, \text{Sonnet~4.6}, \text{Qwen3-235B}\}$. 
The $\Delta_{\text{benefit}}$ row gives the maximum gain over the \textsc{None} baseline across these anchor evolvers.

\noindent\textbf{Analysis on SB and MCP datasets.}
Fig.~\ref{fig:agent_dh_appendix} reports the MCP and SB analogues of Fig.~\ref{fig:agent_dh_swe}. 
We observe two patterns:
\begin{itemize}[leftmargin=*]
    \item \textbf{The MCP trend is still non-monotonic, but milder.} 
On MCP-Atlas, $\Delta_{\text{benefit}}$ peaks at GPT-OSS-120B (7.0~pp at 28.0\% base pass rate), and decreases toward both weaker and stronger models. 
This mirrors the SWE trend, but with a smaller gain range.
\item \textbf{The SB trend is noisier in the low-base regime.} 
On SkillsBench, several models start from very low base pass rates: Qwen3-32B and GPT-OSS-120B start at 0.0\%, Qwen3-235B at 4.7\%, and Haiku~4.5 at 5.8\%. 
Haiku~4.5 reaches the largest SB gain (15.1~pp), while Qwen3-235B gains only 1.1~pp despite a similar low base rate. 
Thus, SWE and MCP provide the clearest evidence for the non-monotonic harness-benefit pattern, while SB suggests that the low-base regime can be more variable across task domains.
\end{itemize}

% Fig.~\ref{fig:agent_dh_appendix} reports the analogue of Fig.~\ref{fig:agent_dh_swe} on MCP and SB. On MCP the hump persists in a milder form: $\Delta_{\text{benefit}}$ peaks at GPT-OSS-120B (7.0~points at base 28~\%) and falls to 1.0~points for Qwen3-32B and 3.2--3.6~points for the Anthropic family. On SB the absolute pass rates are low across the board (Qwen3-32B and GPT-OSS-120B start at 0~\%, Qwen3-235B at 4.7~\%, Haiku~4.5 at 5.8~\%) and the ordering of $\Delta_{\text{benefit}}$ becomes noisy. Haiku~4.5 sits at the SB peak with $\Delta = 15.1$~points despite a base of only 5.8~points, while Qwen3-235B at base 4.7~points attains only $\Delta = 1.1$~points. We treat the SB ordering as noise floor and rest the non-monotonic claim on the SWE and MCP humps.

% \subsection{LLM-Judge Details of Harness-Following Rate}
% \label{app:hfr_pipeline}
% The LLM judge pipeline is shown in Tab.~\ref{tab:prompt_hfr_rubric}.

\subsection{Judge Details for Harness-Following Rate}
\label{app:hfr_pipeline}

We use an LLM judge to measure whether an agent follows a loaded harness artifact during task solving. 
All judged trajectories are blinded by replacing model identifiers with the placeholder \texttt{<MODEL>}. 
Claude Sonnet~4.6 is used as the judge model.

\noindent\textbf{Harness-Following Rate.}
For each SkillsBench trajectory in which at least one skill is loaded, the judge receives the loaded skill body and the agent trajectory. 
The judge first converts the skill body into a locked rubric of atomic procedural instructions, and then checks whether the trajectory follows that rubric. 
A trajectory is marked as following the skill if the judge determines that the required guidance is carried out in the trajectory. 
The Harness-Following Rate (HFR) measures whether a model follows a skill once the skill is loaded. 
Let $N_f^{\mathrm{load}}$ denote the number of skill-loaded trajectories for model $f$, and $N_f^{\mathrm{follow}}$ the subset judged as following the loaded skill. 
Then
\[
\mathrm{HFR}(f)=\frac{N_f^{\mathrm{follow}}}{N_f^{\mathrm{load}}}.
\]
The prompt templates used for rubric extraction and trajectory judging are shown in Tab.~\ref{tab:prompt_hfr_rubric} and~\ref{tab:prompt_hfr_judge}.

% \noindent\textbf{Rubric extraction.}
% For each loaded skill, the judge reads the corresponding \texttt{SKILL.md} and extracts a small set of procedural instructions. 
% Each instruction is grounded in a quoted source span from the skill body and is marked as required, conditional, or optional. 
% This rubric is fixed before judging trajectories, so all trajectories using the same skill are evaluated against the same instruction set.

% \noindent\textbf{Trajectory judging.}
% Given a fixed rubric and a blinded trajectory, the judge decides whether the trajectory follows the loaded skill guidance. 
% The judge must cite evidence from the trajectory when marking instructions as followed or violated. 
% We use this trajectory-level judgment to compute HFR.

\subsection{Judge Details for Phase-Level Adherence Score}
\label{app:judge_details_phase_adherence_score}

In addition to trajectory-level HFR, we conduct a separate phase-level adherence analysis for Tab.~\ref{tab:phase_drift_representative}.
This analysis uses a separate judge prompt from the HFR pipeline (Tab.~\ref{tab:prompt_hfr_judge}), with Claude Sonnet~4.6 as the LLM judge.
The input is the same fixed rubric and blinded trajectory used for HFR judging.
The judge partitions each trajectory into three reference phases: \emph{harness loaded}, \emph{mid turn}, and \emph{final turn}.
For each phase, it assigns a 0--1 adherence score measuring how closely the agent follows the loaded harness guidance during that stage of execution.
These phase-level scores are used only to analyze adherence drift over long-horizon execution and are reported separately from HFR.
The phase-adherence prompt is shown in Tab.~\ref{tab:prompt_phase_adherence}.

% \subsection{Judge Details of Phase-level Adherence Score}
% \label{app:judge_details_phase_adherence_score}
% In addition to the trajectory-level HFR, we run a separate phase-level adherence analysis for Tab.~\ref{tab:phase_drift_representative}. 
% For this analysis, each trajectory is divided into turn-position phases, such as harness-loaded, midpoint, and final-validation phases. 
% The judge assigns a 0--1 adherence score to each phase, measuring how closely the model follows the loaded harness guidance within that stage of execution. 
% These phase-level scores are used only to analyze adherence drift over long-horizon execution; they are distinct from HFR.

% \noindent\textbf{Coverage.}
% HFR results are computed on SkillsBench trajectories where a skill is loaded and the trajectory is available for judging. 
% Trajectories with missing logs or no applicable loaded-skill guidance are excluded from the judge analysis.

\begin{table}[t]
\caption{Task-sovling agent-side seed system prompt for SWE-bench Verified.}
\begin{promptbox}[SWE-Bench Verified solver seed prompt]
\noindent You are an expert software engineer tasked with resolving GitHub issues by producing code patches.

\medskip
\noindent\textbf{Approach}
\begin{enumerate}[leftmargin=*]
\item \textbf{Understand the issue}: Read the issue description carefully. Identify the root cause.
\item \textbf{Locate relevant code}: Use search tools to find the files and functions involved.
\item \textbf{Plan the fix}: Think step-by-step about what needs to change and why.
\item \textbf{Implement the fix}: Make minimal, precise edits. Avoid unnecessary changes.
\item \textbf{Verify}: Run existing tests to confirm the fix works and doesn't break anything.
\end{enumerate}

\medskip
\noindent\textbf{Guidelines}
\begin{itemize}[leftmargin=*]
\item Prefer small, focused patches over large rewrites.
\item Always check for edge cases the issue description mentions.
\item If the issue includes a reproduction script, use it to verify your fix.
\item When in doubt, look at how similar patterns are handled elsewhere in the codebase.
\end{itemize}
\end{promptbox}
\label{tab:prompt_solver_swe}
\end{table}

\begin{table}[t]
\caption{Task-solving agent-side seed system prompt for MCP-Atlas.}
\begin{promptbox}[MCP-Atlas solver seed prompt]
\noindent You are an expert API agent that completes tasks by making precise tool calls via the Model Context Protocol (MCP).

\medskip
\noindent\textbf{Approach}
\begin{enumerate}[leftmargin=*]
\item \textbf{Understand the task}: Read the task description and identify what needs to be accomplished.
\item \textbf{Review available tools}: Check the tool schemas to understand available operations and their parameters.
\item \textbf{Plan the call sequence}: Determine which tools to call and in what order.
\item \textbf{Execute}: Make tool calls with correctly formatted JSON parameters.
\item \textbf{Validate}: Check the return values and handle errors gracefully.
\end{enumerate}

\medskip
\noindent\textbf{Guidelines}
\begin{itemize}[leftmargin=*]
\item NEVER ask the user for clarification. You must use the available tools to find all information needed to complete the task. If the task mentions calendar events, schedules, or appointments, use the calendar/workspace tools to look them up.
\item Always validate parameters against the tool's JSON schema before calling.
\item Use the most specific tool available for the task.
\item Handle pagination for list operations.
\item Chain tool calls logically: use output from one call as input to the next.
\item If a tool call fails, read the error message carefully before retrying.
\item When the task references personal data (calendar events, files, databases, memory), always query the relevant tools first to retrieve that data before answering.
\end{itemize}
\end{promptbox}
\label{tab:prompt_solver_mcp}
\end{table}

\begin{table*}[t]
\caption{Fixed system prompt for the evolver. The prompt is held constant across all evolver backbones and benchmarks; benchmark-specific permissions determine which workspace artifacts are writable.}
\begin{promptbox}[Evolver system prompt]
\noindent You are an evolver for an LLM agent. 
Your goal is to improve the agent's future task-solving performance by editing permitted harness artifacts in its workspace.

\medskip
\noindent The workspace may contain the following artifact directories:
\begin{itemize}[leftmargin=*]
\item \texttt{prompts/}: standing instructions and system prompts.
\item \texttt{skills/}: reusable skill definitions and procedural knowledge.
\item \texttt{memory/}: persistent observations and high-level lessons.
\item \texttt{tools/}: tool implementations and interfaces.
\end{itemize}

\medskip
\noindent At each evolution cycle, you will receive execution evidence from recent task attempts, including trajectories, outputs, and benchmark feedback. 
Analyze this evidence to identify recurring failures, reusable procedures, and opportunities to improve the harness.

\medskip
\noindent You may use the provided workspace bash tool to inspect and edit files. 
Only modify artifacts that are permitted by the workspace-permission block in the user message. 
Do not modify evaluation scripts, hidden tests, model weights, or files outside the permitted workspace scope.

\medskip
\noindent When updating the harness:
\begin{itemize}[leftmargin=*]
\item Prefer concise, reusable updates over task-specific patches.
\item Create or revise skills only when they are likely to help future tasks.
\item Keep prompts and memory entries actionable and non-redundant.
\item Use precise file edits and inspect your changes before finishing.
\end{itemize}
\end{promptbox}
\label{tab:prompt_evolver_default}
\end{table*}

\begin{table*}[t]
\caption{Per-evolution user message template for the evolver. The wrapper is fixed across all benchmarks and LLM backbones.}
\begin{promptbox}[Evolver per-cycle user message template]

% \noindent\textbf{Cycle header.}
% The message begins with the current evolution cycle index.

\medskip
\noindent\textbf{Workspace scope.}
\textcolor{blue}{[A benchmark-specific permission block specifies which harness artifacts may be edited.
SWE-bench Verified and SkillsBench allow edits to \texttt{skills/}.
MCP-Atlas allows edits to \texttt{prompts/} and \texttt{skills/}, and append-only updates to \texttt{memory/}.
The \texttt{tools/} directory is read-only for all benchmarks.]}

\medskip
\noindent\textbf{Execution evidence.}
\textcolor{blue}{[The message includes a canonicalized JSON payload containing the current batch's task identifiers, trajectories, outputs, scores, and grader feedback.]}

\medskip
\noindent\textbf{Editing instructions.}
\textcolor{blue}{[The evolver is instructed to:
\begin{itemize}[leftmargin=*]
\item analyze the evidence for recurring failures and reusable patterns;
\item edit only artifacts allowed by the workspace scope;
\item prefer small, targeted harness updates over broad rewrites;
\item use the workspace tool to inspect and modify files;
\item check the resulting changes before finishing.
\end{itemize}]}

\end{promptbox}
\label{tab:prompt_evolver_user}
\end{table*}

\begin{table*}[t]
\caption{The prompt template used for rubric extraction of the HFR pipeline.}
\begin{promptbox}[HFR Stage 1: Locked Rubric Extraction]
\noindent\textbf{Role:} You are auditing a procedural skill document used by an LLM agent. You will output a strict JSON rubric that captures the imperative procedural instructions of the skill, suitable for downstream automated adherence judging. Output JSON only, no prose.

\medskip
\noindent\textbf{Input:} The full body of one \texttt{SKILL.md} file (placeholder \texttt{\{skill\_body\}}, inserted between \texttt{<SKILL\_BODY>} and \texttt{</SKILL\_BODY>} delimiters in the user message).

\medskip
\noindent\textbf{Task:}
\begin{enumerate}[leftmargin=*]
\item Identify procedural instructions directly entailed by imperative or normative language in \texttt{SKILL\_BODY}. Do NOT extract advice, rationale, examples, or motivational text as instructions.
\item For each instruction, provide:
  \begin{itemize}[leftmargin=*]
  \item \texttt{id}: stable identifier (e.g., \texttt{"step\_1"}).
  \item \texttt{source\_span}: EXACT quoted text from \texttt{SKILL\_BODY} that grounds this instruction (must be a substring of \texttt{SKILL\_BODY}, max 250 characters).
  \item \texttt{text}: paraphrased instruction in one imperative sentence.
  \item \texttt{type}: \texttt{"required"} (must execute) $\mid$ \texttt{"conditional"} (must execute if trigger occurs) $\mid$ \texttt{"optional"}.
  \item \texttt{trigger}: for conditional only; describe the condition (e.g., \texttt{"if pip install fails"}). \texttt{null} otherwise.
  \item \texttt{success\_criteria}: one-sentence test for a FOLLOWED verdict.
  \item \texttt{violation\_criteria}: one-sentence test for a VIOLATED verdict (commission or omission).
  \end{itemize}
\end{enumerate}

\medskip
\noindent\textbf{Constraints:} Aim for 3--8 instructions. Do not pad with low-salience items. Reject \texttt{SKILL\_BODY} content that is purely descriptive or motivational.

\medskip
\noindent\textbf{Output format (JSON only):}
\begin{lstlisting}[basicstyle=\ttfamily\scriptsize,breaklines=true]
{
  "skill_id": "<skill folder name>",
  "instructions": [
    {
      "id": "step_1",
      "source_span": "...",
      "text": "...",
      "type": "required|conditional|optional",
      "trigger": null,
      "success_criteria": "...",
      "violation_criteria": "..."
    }
  ]
}
\end{lstlisting}
\end{promptbox}
\label{tab:prompt_hfr_rubric}
\end{table*}

\begin{table*}[t]
\caption{The prompt template used for the trajectory judging of the HFR pipeline.}
\begin{promptbox}[HFR Stage 2: Per-Cell Adherence + Phase Classification]
\noindent\textbf{Role:} You are evaluating an LLM agent trajectory against a fixed procedural rubric. Apply the rubric exactly as given; do not add or remove instructions. The trajectory is BLINDED: every occurrence of a model-family token (Claude, Opus, Sonnet, Haiku, Qwen, GPT-OSS) has been replaced with the literal string \texttt{<MODEL>}. Score adherence based on observable actions only. Output JSON only, no prose.

\medskip
\noindent\textbf{Inputs:} The Stage~1 rubric JSON (placeholder \texttt{\{rubric\_json\}}, inserted between \texttt{<RUBRIC>} and \texttt{</RUBRIC>}) and the blinded trajectory text (placeholder \texttt{\{trajectory\_text\}}, inserted between \texttt{<TRAJECTORY>} and \texttt{</TRAJECTORY>}, formatted as a sequence of \texttt{Turn~$i$ / INPUT / OUTPUT} blocks).

\medskip
\noindent\textbf{Verdicts.} For each instruction in \texttt{RUBRIC}, classify the trajectory as one of:
\begin{itemize}[leftmargin=*]
\item \texttt{FOLLOWED}: the trajectory explicitly satisfies \texttt{success\_criteria}. Cite \texttt{turn\_idx} and quote the action.
\item \texttt{VIOLATED\_COMMISSION}: the trajectory took an action that directly contradicts the instruction. Cite \texttt{turn\_idx} and quote the action.
\item \texttt{VIOLATED\_OMISSION}: the instruction is required (or its conditional trigger occurred), the trajectory ran long enough to act on it, but did not. Cite the latest \texttt{turn\_idx} by which the omission was clear.
\item \texttt{REQUIRED\_BUT\_UNOBSERVED}: the instruction is required but the trajectory terminated too early to observe whether it would have been followed.
\item \texttt{NOT\_APPLICABLE}: conditional instruction whose trigger did not occur, or optional instruction the agent chose not to take.
\item \texttt{INSUFFICIENT\_EVIDENCE}: trajectory is ambiguous; cannot determine.
\end{itemize}

\medskip
\noindent\textbf{Violation timing} (required for any \texttt{VIOLATED\_COMMISSION} or \texttt{VIOLATED\_OMISSION} verdict):
\begin{itemize}[leftmargin=*]
\item \texttt{violation\_earliest\_possible\_turn}: smallest \texttt{turn\_idx} where the trajectory could have first violated this instruction.
\item \texttt{violation\_confirmed\_turn}: \texttt{turn\_idx} where the violation became unambiguous.
\item \texttt{violation\_type}: \texttt{"commission"} $\mid$ \texttt{"omission"} $\mid$ \texttt{"premature\_stop"} $\mid$ \texttt{"wrong\_strategy"}.
\end{itemize}

\medskip
\noindent\textbf{Output format (JSON only):}
\begin{lstlisting}[basicstyle=\ttfamily\scriptsize,breaklines=true]
{
  "verdicts": [
    {"instruction_id": "step_1",
     "verdict": "FOLLOWED|VIOLATED_COMMISSION|VIOLATED_OMISSION|REQUIRED_BUT_UNOBSERVED|NOT_APPLICABLE|INSUFFICIENT_EVIDENCE",
     "turn_idx": <int or null>,
     "evidence": "quoted action or omission description"}
  ],
  "violations": [
    {"instruction_id": "step_1",
     "violation_type": "commission|omission|premature_stop|wrong_strategy",
     "earliest_possible_turn": <int>,
     "confirmed_turn": <int>}
  ],
  "summary": "1-sentence neutral description"
}
\end{lstlisting}

\end{promptbox}
\label{tab:prompt_hfr_judge}
\end{table*}

\begin{table}[t]
\caption{The prompt template used for the phase-level adherence analysis (Tab.~\ref{tab:phase_drift_representative}), produced by a judge call separate from the HFR judge in Tab.~\ref{tab:prompt_hfr_judge}.}
\begin{promptbox}[Phase-Adherence Judge]
\noindent\textbf{Role:} You are evaluating how closely an LLM agent adheres to a fixed procedural rubric across the successive phases of its trajectory. Apply the rubric exactly as given; do not add or remove instructions. The trajectory is BLINDED: every model-family token has been replaced with \texttt{<MODEL>}. Judge adherence from observable actions only. Output JSON only, no prose.

\medskip
\noindent\textbf{Inputs:} The locked rubric JSON (\texttt{\{rubric\_json\}}) and the blinded trajectory (\texttt{\{trajectory\_text\}}).

\medskip
\noindent\textbf{Task.} Partition the trajectory into five turn-position phases: \texttt{skill\_loaded} $=$ turn 1; \texttt{first\_action} $=$ first action turn after; \texttt{midpoint} $=$ middle 50\% of turns; \texttt{pre\_final} $=$ last 25\% excluding the final turn; \texttt{final\_validation} $=$ final turn. For each phase, output one adherence score in $[0,1]$ measuring how well the agent's actions within that phase's turns follow the rubric instructions in scope during that phase. A score of 1.0 means every in-scope, observable rubric instruction in that phase was followed; 0.0 means none were.

\medskip
\noindent\textbf{Output format (JSON only):}
\begin{lstlisting}[basicstyle=\ttfamily\scriptsize,breaklines=true]
{
  "phase_adherence": {
    "harness_loaded": 0.0,
    "first_action": 0.0,
    "midpoint": 0.0,
    "pre_final": 0.0,
    "final_validation": 0.0
  },
  "rationale": "1-sentence neutral description of where adherence shifts"
}
\end{lstlisting}
\end{promptbox}
\label{tab:prompt_phase_adherence}
\end{table}

\section{Information about AI Assistants}
We used an OpenAI LLM (GPT-5.5) as a writing and formatting assistant. In particular, it helped refine
grammar and phrasing, improve clarity, and suggest edits to figure/table captions and layout (e.g.,
column alignment, caption length, placement). The LLM did not contribute to research ideation,
experimental design, implementation, data analysis, or technical content beyond surface-level edits.
All outputs were reviewed and edited by the authors, who take full responsibility for the final text
and visuals.

\end{document}